\RequirePackage{fix-cm}
\documentclass[twocolumn, times]{svjour3}       
\smartqed  
\usepackage{cite}
\usepackage{color}
\usepackage{times}
\usepackage{epsfig}
\usepackage{graphicx}
\usepackage{graphics}
\usepackage{amsmath}
\usepackage{amssymb}
\usepackage{multirow}
\usepackage{mathrsfs}
\usepackage{rotating}
\usepackage{bbm}
\usepackage{verbatim}
\usepackage{array}
\usepackage{booktabs}
\usepackage{tabularx}
\usepackage{url}
\usepackage{marvosym}
\usepackage{natbib}
\usepackage[colorlinks,linkcolor=red,anchorcolor=blue,citecolor=black,CJKbookmarks=True]{hyperref}
\hypersetup{
    colorlinks=true,
    urlcolor=black
}
\usepackage[switch]{lineno}

\def\eg{\emph{e.g.}}

\def\etal{\emph{et al.}}
\def\ie{\emph{i.e.}}

\begin{document}

\title{Instance Segmentation in the Dark
}


\author{Linwei~Chen$^{1,2}$ \and
        Ying~Fu$^{2}$ \and
        Kaixuan~Wei$^{4}$ \and
        Dezhi~Zheng$^{1}$ \and
        Felix~Heide$^{3,5}$
}



\institute{
$\textrm{\Letter}$ Ying Fu\\
\hspace*{3mm} fuying@bit.edu.cn\\ \\
$1$ Advanced Research Institute of Multidisciplinary Science, Beijing Institute of Technology, Beijing, China\\ \\
$2$ School of Computer Science and Technology, Beijing Institute of Technology, Beijing, China\\ \\
$3$ Department of Computer Science, Princeton University, Princeton, New Jersey, USA \\ \\
$4$ Department of Electrical and Computer Engineering, McGill University, Quebec, Canada \\ \\
$5$ Algolux, Rue Richardson, Montreal, Canada \\ \\
}

\maketitle

\begin{abstract}
Existing instance segmentation techniques are primarily tailored for high-visibility inputs, but their performance significantly deteriorates in extremely low-light environments. 
In this work, we take a deep look at instance segmentation in the dark and introduce several techniques that substantially boost the low-light inference accuracy.
The proposed method is motivated by the observation that noise in low-light images introduces high-frequency disturbances to the feature maps of neural networks, thereby significantly degrading performance.
To suppress this ``feature noise", we propose a novel learning method that relies on an adaptive weighted downsampling layer, a smooth-oriented convolutional block, and disturbance suppression learning.
These components effectively reduce feature noise during downsampling and convolution operations, enabling the model to learn disturbance-invariant features. 
Furthermore, we discover that high-bit-depth RAW images can better preserve richer scene information in low-light conditions compared to typical camera sRGB outputs, thus supporting the use of RAW-input algorithms. 
Our analysis indicates that high bit-depth can be critical for low-light instance segmentation.
To mitigate the scarcity of annotated RAW datasets, we leverage a low-light RAW synthetic pipeline to generate realistic low-light data.
In addition, to facilitate further research in this direction, we capture a real-world low-light instance segmentation dataset comprising over two thousand paired low/normal-light images with instance-level pixel-wise annotations. 
Remarkably, without any image preprocessing, we achieve satisfactory performance on instance segmentation in very low light (4~\% AP higher than state-of-the-art competitors), meanwhile opening new opportunities for future research.
Our code and dataset are publicly available to the community\footnote{\url{https://github.com/Linwei-Chen/LIS}}.

\end{abstract}

\section{Introduction}
\label{sec:intro}
Instance segmentation, as a technique that solves the problem of object detection and semantic segmentation at the instance level simultaneously, plays a critical role in helping computers understand visual information and thus supports applications such as robotics \citep{fang2018multi, mohan2021efficientps} and autonomous driving \citep{de2017autonomous}, and etc. 

\begin{figure}
\centering
\setlength\tabcolsep{1pt}
\small
\scalebox{0.8}{
\begin{tabular}{cccc}
\includegraphics[width=0.6\linewidth]{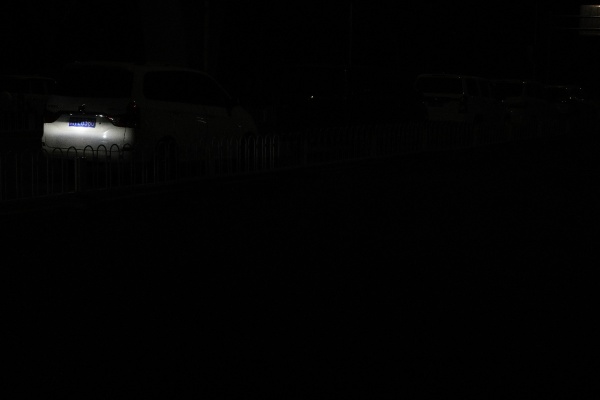}
&\includegraphics[width=0.6\linewidth]{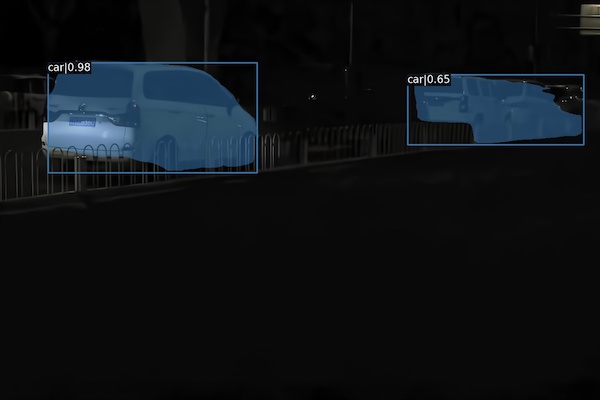}
\\
(a) Short-exposure camera output  &   (b) Result of enhanced image 
\\
\includegraphics[width=0.6\linewidth]{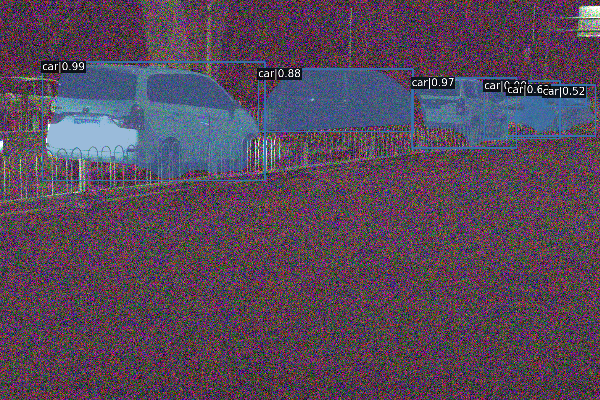}
&\includegraphics[width=0.6\linewidth]{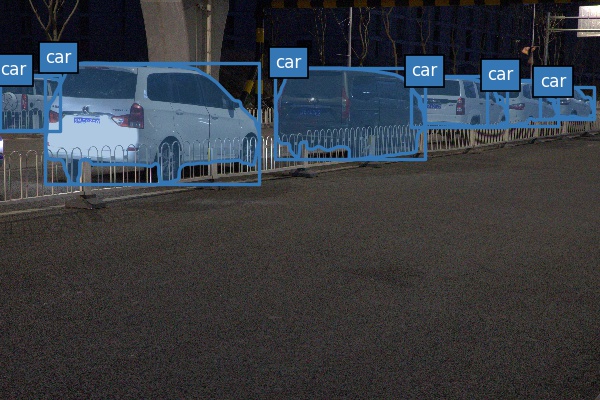}
\\
(c) Our result from RAW data   &  (d) Long-exposure reference 
\end{tabular}
}
\vspace{-1mm}
\caption{\small Extreme low-light instance segmentation with Mask R-CNN \citep{MaskRCNN2017} based upon 
(a) a short-exposure low-light image directly; 
(b) the image preprocessed by the state-of-the-art  low-light enhancement algorithm \citep{guo2020zero} plus the denoising algorithm \citep{gu2019self};
(c) our proposed method on the amplified RAW image (the displayed result here is converted from RAW space to sRGB for visualization); and (d) the corresponding long-exposure reference image. 
}
\label{fig:example}
\vspace{-3mm}
\end{figure}

With the advent of deep learning, many learned instance segmentation methods have been proposed \citep{MaskRCNN2017, yolact2019, htc2019, centermask2020, blendmask2020}, and have achieved promising performance in well-lit scenarios. 
However, these methods often fail to work well in dimly-lit environments, where the detailed contents are ``buried" by severe noise caused by limited photon count and imperfections in photodetectors.
While low-light instance segmentation is an important task, there are few methods or datasets specifically designed for this purpose. Relevant low-light recognition/detection methods~\citep{2020yolodark, 2021meat} and datasets~\citep{2019exdark, 2021nod, 2021Crafting} are still in their infancy.
In this context, a common and simple solution is to combine image enhancement/denoising algorithms with instance segmentation models~\citep{2020connecting}. However, the additional image restoration process increases the computational cost and the overall latency of the pipeline. 
Even then, under extremely low light, these image restoration algorithms, as shown in Figure~\ref{fig:example}, can only recover limited scene information due to the permanent loss of image details in typical camera sRGB outputs.

\begin{figure*}[!t]
\centering
\setlength\tabcolsep{1pt}
\small
\scalebox{0.80}{
\begin{tabular}{cccc}
\includegraphics[width=0.6\linewidth]{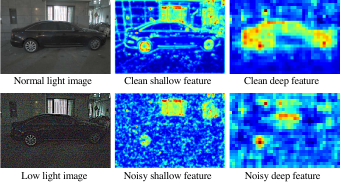} \hspace{+2mm}
&\includegraphics[width=0.6\linewidth]{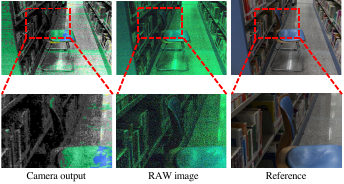}
\\
(a) Degraded feature maps under low light 
& (b) Amplified camera sRGB output and the corresponding RAW image 
\end{tabular}
}
\caption{
Illustration of our key observations under dark regimes that drive our method design:
(a) Degraded feature maps under low light. 
For clean normal-light images, the instance segmentation network is able to clearly capture the low-level (\eg, edges) and high-level (\ie, semantic responses) features of objects in shallow and deep layers, respectively.  
However, for noisy low-light images, shallow features can be corrupted and full of noise, and the deep features show lower semantic responses to objects.
(b) Comparison between camera sRGB output and RAW image in the dark. 
Due to significantly low SNR, the 8-bit camera output loses much of the scene information, for example, the seat backrest structure is barely discernible, whereas is still recognizable in the RAW counterpart \textbf{(Zoom in for better details)}.
}
\label{fig:observation}
\end{figure*}

In this work, we aim to craft a practical low-light instance segmentation framework in an end-to-end manner with marginal additional computational cost.
To this end, we look deep into the instance segmentation model and analyze how low-light images {\color{black}harm} the instance segmentation performance.
We observe that the noise in low-light images brings ``feature noise" (\emph{i.e.}, high-frequency disturbance as shown in Figure~\ref{fig:observation}{\color{red}(a)}) into features inside the neural network.
This leads to lower semantic responses of scene content in deep feature maps, therefore causing the low recall of scene content and degenerating the performance.
This important phenomenon is also observed in adversarial defense/attack literatures~\citep{2013intriguing, 2019featuredenoising}, which suggests that restoring features from samples with adversarial noise can be critical for model {\color{black}robustness}~\citep{2019featuredenoising}.
Motivated by this observation, we propose to augment existing instance segmentation methods with an adaptive weight downsampling layer, smooth-oriented convolutional block and disturbance suppression learning. 
They substantially improve the capability of models to learn noise-resisted features and thus boost the low-light segmentation accuracy appreciably. 
{\color{black}
It is worth noting that they are model-agnostic and lightweight or even cost-free.
}

{\color{black}
Specifically, the adaptive weight downsampling layer can generate content-aware low-pass filters during feature map downsampling. 
It aggregates local features adaptively and suppresses the high-frequency disturbance caused by noise as well as keeping the details in deep features. 
The smooth-oriented convolutional block enhances the ordinary convolutional layers by adding a smooth-oriented convolution branch.
It helps to improve the robustness of the network for feature noise and can be re-parameterized~\citep{ding2021repvgg} to the normal convolutional layer.
The disturbance suppression learning guides networks to learn noise-resisted features, so as to keep stable semantic responses of scene content for noisy low-light images.
Remarkably, they are model-agnostic.
And only minor computational overhead is added by the adaptive weight downsampling layer, while smooth-oriented convolutional block and disturbance suppression learning introduce no extra computational cost since they are only involved during training.  
}

Moreover, we notice that the high bit-depth can be crucial for low-light conditions. 
Thus to reduce the loss of scene information in dark conditions, instead of 8-bit sRGB camera outputs, we use 14-bit RAW sensor data as inputs, which have higher bit-depth and better potential to preserve scene information even under extreme low-light conditions (See Figure~\ref{fig:observation}{\color{red}(b)}).
However, to date, there is no low-light RAW image dataset for instance segmentation, and its collection and annotation could be tremendously labor-intensive.
To solve this, we leverage a low-light RAW synthetic pipeline. 
It can generate realistic RAW image datasets from any existing sRGB image datasets (\eg, PASCAL VOC 2012~\citep{2010pascal}, COCO~\citep{mscoco2014}), which makes an end-to-end training of RAW-input instance segmentation model feasible.

To systematically examine the performance of existing approaches under real low-light environments, we also capture and label a low-light instance segmentation (LIS) dataset with 2230 pairs of low/normal-light images, covering diverse real-world indoor/outdoor low-light scenes.  
Extensive experiments validate the superior instance segmentation performance of our method in the dark, consistently outperforming existing methods in accuracy and computation cost.

Our main contributions can be summarized as follows:
\begin{itemize}
\item We propose an adaptive weighted downsampling layer, smooth-oriented convolutional block and disturbance suppression learning to address the high-frequency disturbance within deep features that occurred in very low light. Interestingly, they also benefit the normal-lit instance segmentation.
\item We exploit the potentials of RAW-input design for low-light instance segmentation and leverage a low-light RAW synthetic pipeline to generate realistic low-light RAW images from existing datasets, which facilitates end-to-end training.
\item We collect a real-world low-light dataset with precise pixel-wise instance-level annotations, namely LIS, which covers more than two thousand scenes and can serve as a benchmark for instance segmentation in the dark.
On LIS, our approach outperforms state-of-the-art competitors in terms of both segmentation accuracy and inference speed by a large margin.
\end{itemize}

\begin{figure}[t]
\centering
\includegraphics[width=0.98\linewidth]{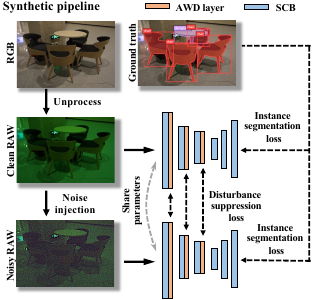} 
\caption{
Overview of our proposed method.
The adaptive weighted downsampling (AWD) layer, smooth-oriented convolutional block (SCB), and disturbance suppression loss are designed to reduce the feature disturbance caused by noise, and the low-light RAW synthetic pipeline is employed to facilitate end-to-end training of instance segmentation on RAW images.
}
\label{fig:overview}
\end{figure}


\section{Related Work}
\label{sec:related}
{\color{black}
}

\noindent\textbf{Normal instance segmentation.~}
With the birth of deep learning, the field of computer vision has flourished~\citep{2021coded, wei2021physics, zhang2022guided}.
Instance segmentation~\citep{MaskRCNN2017, htc2019, yolact2019, blendmask2020, centermask2020, chen2022hybrid, chen2021efficient} aim to predict the class label and the pixel-specific instance mask for objects. 
It localizes different classes of object instances present in various images. 
Many methods~\citep{MaskRCNN2017, htc2019} rely on Faster R-CNN \citep{fasterRCNN2015} detector.
By adopting the detect-then-refine strategy, they can achieve superior performance but run at a relatively slow speed.
Other methods~\citep{yolact2019, blendmask2020, centermask2020} are based on simple yet effective detectors~\citep{2016yolo, retinanet2017,fcos2019}, they can run in real-time and achieve competitive accuracy.
Though achieving significant progress, most existing works only consider normal-light scenarios and largely overlooked low-light conditions.

\vspace{+1mm}
\noindent\textbf{Instance segmentation in the dark.~}
To adopt instance segmentation for very low-light, a straightforward solution is casting the low-light enhancement methods~\citep{2018sid,jiang2021enlightengan,yang2020advancing, zhang2021beyond, lv2021attention} or image denoising methods~\citep{tan2007multivariate,hajiaboli2011anisotropic,gu2019self,hahn2011orientation,ulyanov2020deep, 2021invertible} 
as pre-processing steps.
Compared with normal-light instance segmentation, research for low-light instance segmentation is at its early stage and relatively less at present.

A diverse body of work explores low-light classification.
Gnanasambandam \etal~\citep{2020qis} present a new low-light image classification method using Quanta Image Sensors (QIS) and show promising results by utilizing a student-teacher learning scheme to classify the noisy QIS raw data.

As for low-light object detection, Liu \etal~\citep{2020connecting} use a high-level vision model to guide the training of denoiser and demonstrate the benefit for image denoising and high-level vision tasks. 
Diamond \etal~\citep{2021dirty} introduce Anscombe networks, which are lightweight neural camera ISP for demosaicking and denoising.
It shows desirable performance on low-light classification by jointly learning Anscombe networks with classification networks.
Julca-Aguilar \etal~\citep{2021gated3d} propose a novel 3D object detection modality that exploits temporal illumination cues from a low-cost monocular gated imager.
It shows better potential to deal with low-light or low-contrast regions.
Wang \etal~\citep{2021hla} propose a joint High-Low Adaptation (HLA) framework. 
By adopting a bidirectional low-level adaptation and multi-task high-level adaptation scheme, the proposed HLA-Face outperforms state-of-the-art methods even without using dark face labels for training.
Sasagawa \etal~\citep{2020yolodark} propose glue layer to ``glue'' SID model~\citep{2018sid} and YOLO model~\citep{2016yolo} together. 
Cui \etal~\citep{2021meat} propose to learn the intrinsic visual structure by encoding and decoding the realistic illumination-degrading transformation. 
They achieve desired performance on the low-light classification or low-light objection detection task but do not consider more challenging low-light instance segmentation.

{\color{black}
\vspace{+1mm}
\noindent\textbf{Low-light synthesis.~}
The low-light enhancement methods usually need low-light/normal-light image pairs for training~\citep{lore2017llnet, wei2018deep, wang2018gladnet, 2018sid, lamba2021restoring, 2019effectiveConvLSTM, 2021temporalconsistency}, which is hard to obtain.
Some works~\citep{guo2020zero, jiang2021enlightengan, 2022legan} solve it by learning in a zero-reference way or utilizing unpaired images for training. 
And some works explore synthesizing low-light images from normal-light images.
Retinex-Net~\citep{wei2018deep} collects normal-light RAW images from RAISE~\citep{2015raise} and makes their histogram of Y channel in YCbCr fit the result in low-light images from public datasets, thus getting synthetic low-light images with Adobe Lightroom. 
The GLADNet~\citep{wang2018gladnet} also synthesizes low-light images from RAW images in RAISE~\citep{2015raise}, which is done by adjusting the exposure, vibrance, and contrast parameters.
And the recent works~\citep{xu2020learning,punnappurath2022day} make progress in synthesizing low-light images by taking noise into consideration, but they still rely on RAW images that existing datasets for instance segmentation do not have. 
Though some works~\citep{lore2017llnet, 2021meat} try to synthesize low-light sRGB images from normal-light ones, they only consider simple Gaussian and Poisson noise. 
Moreover, they are not applicable for synthesizing low-light RAW images from existing sRGB datasets. 
To solve this, we leverage unprocessing~\citep{brooks2019unprocessing} and employ a recently proposed physics-based noise model~\citep{wei2020physics, wei2021physics} to synthesize realistic low-light RAW images from any sRGB images with labels.
}

\vspace{+1mm}
\noindent\textbf{Datasets for low-light instance segmentation.~}
Existing common datasets for instance segmentation, \eg, PASCAL VOC~\citep{2010pascal}, cityscapes~\citep{cityscapes2016} and COCO~\citep{mscoco2014}, play an important role in the progress of instance segmentation algorithms under normal illumination.
And there are several datasets available for nighttime detection~\citep{2021nod, 2019exdark, 2021Crafting, liu2021benchmarking} and semantic segmentation~\citep{2018darkmodel, 2019darkzurich, 2021nightcity}.
There are also some benchmark studies for understanding poor visibility environments~\citep{2020advancing, 2018foggy, dai2020curriculum, 2022legan}.
However, images in these datasets are captured in somewhat dim environments instead of extremely low-light, whose noise levels are low.
Moreover, they are not suitable for instance segmentation due to the lack of instance-level pixel-wise labels.
To better develop instance segmentation in extremely low-light, we collect and annotate a real-world low-light image dataset with precise pixel-wise instance-level annotations called Low-light Instance Segmentation (LIS).

\section{Learning Segmentation in Low Light}
{\color{black}
The overview of the proposed method is shown in Figure~\ref{fig:overview}.
In this section, we first describe our motivation. 
Then we introduce the low-light RAW synthetic pipeline in Section~\ref{sec:low-light-synthesis}.
Finally, we show the details of the Adaptive Weighted Downsampling (AWD) layer, Smooth-oriented Convolutional Block (SCB), and Disturbance Suppression Learning (DSL) in Sections~\ref{sec:awd}, \ref{sec:scb} and \ref{sec:dsl}, respectively. 
}

\subsection{Motivation}
{\color{black}
A practical low-light instance segmentation framework should be accurate and efficient.
We notice the RAW images have better potential to recover scene information (see Figure~\ref{fig:observation} {\color{red}(b)}), owing to higher bit depth.
This should benefit the low-light instance segmentation.
Nevertheless, collecting RAW image dataset for low-light instance segmentation is {\color{black}expensive} and labor-intensive, and it is better if the existing normal-light image datasets~\citep{2010pascal, mscoco2014} could be utilized for training target models. 
To this end, 
we leverage unprocessing and noise injection to synthesize realistic low-light RAW images from any sRGB images with labels.

Furthermore, we observe noise in low-light images can disturb the prediction, and solutions of this degradation usually bring extra computational cost, \eg, prepending enhancing/denoising step.
To avoid this, instead of denoising the image, we aim to denoise the feature inside the instance segmentation model, \emph{i.e.}, suppress the high-frequency disturbance within feature maps (see Figure~\ref{fig:observation} {\color{red}(a)}).
This should be more economical in computation than adding extra image-enhancing/denoising models.
Next, we introduce the proposed method in detail.
}

\begin{figure*}[t!]
\centering
\includegraphics[width=0.98\linewidth]{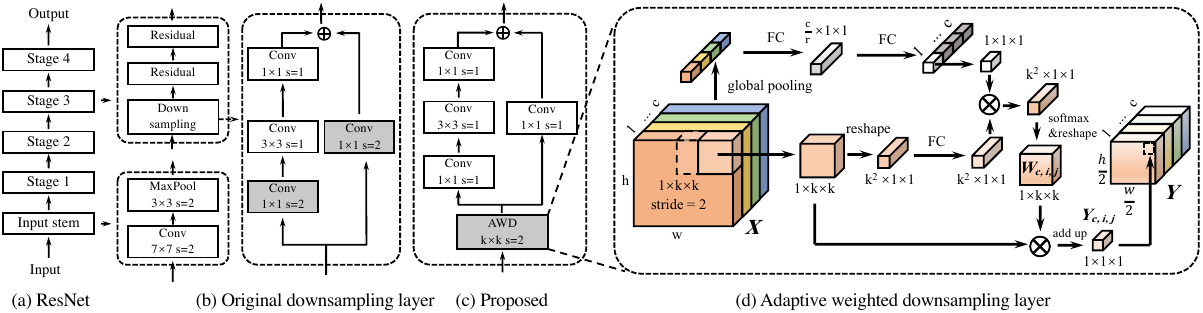}    
\caption{
Illustration of adaptive weighted downsample layer.
The original downsampling operation in ResNet~\citep{resnet2016} is done by convolutional layers (b) with stride$=$2, which fails to leverage 3/4 of spatial features for feature noise suppression.
The proposed adaptive weighted downsampling (AWD) layer (c) can selectively aggregate all surrounding features to generate downsampled features with less feature noise.
FC in (d) indicates a fully connected layer, {\color{black}and ``r'' indicates the channel reduction ratio.}
}
\label{fig:resnet}
\vspace{-4mm} 
\end{figure*}

\begin{figure}[t!]
\centering
\includegraphics[width=0.98\linewidth]{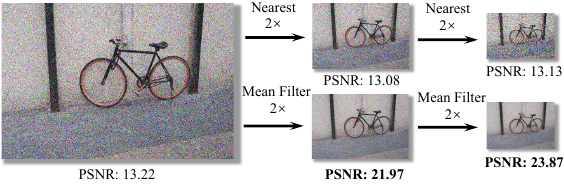}
\caption{
Downsampling the noisy image with low-pass filters, \eg, mean filter, can suppress the noise, whereas nearest neighbor interpolation cannot.}
\label{fig:toy}
\end{figure}
\begin{figure}[t!]
\centering
\includegraphics[width=0.88\linewidth]{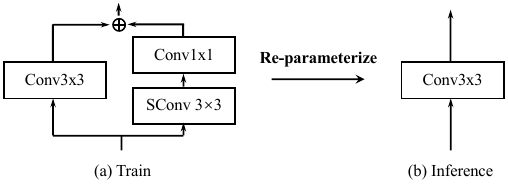}
\caption{
Illustration of the smooth-oriented convolutional block.
SConv indicates the smooth-oriented convolution.
(a) It explicitly employs a linear combination of multiple branches during the training stage.
(b) And in the inference stage, it can be folded back to a normal $3\times3$ convolutional layer by using the re-parameterization technique~\citep{ding2021repvgg}.
}
\label{fig:scb}
\end{figure}

\subsection{Low-light RAW Synthetic Pipeline}
\label{sec:low-light-synthesis}
Our low-light RAW synthetic pipeline consists of two steps, \emph{i.e.}, unproccessing and noise injection. 
We introduce them one by one.

\vspace{+1mm}
\noindent \textbf{Unprocessing.~}
Collecting a large-scale RAW image dataset is {\color{black}expensive} and time-consuming, hence we consider utilizing existing sRGB image datasets~\citep{2010pascal, mscoco2014}.  
The sRGB image is obtained from RAW images by a series of image transformations of on-camera image signal processing (ISP), \eg, tone mapping, gamma correction, color correction, white balance, and demosaicking.
With the help of the unprocessing operation~\citep{brooks2019unprocessing}, we can invert these image processing transformations, and RAW images can be obtained.
In this way, we can create a RAW dataset with zero cost.
	
\vspace{+1mm}
\noindent \textbf{Noise injection.~}
After obtaining clean RAW images by unprocessing, to simulate real noisy low-light images, we need to inject noise into RAW images.
To yield more accurate results for real complex noise, we employ a recently proposed physics-based noise model~\citep{wei2020physics, wei2021physics}, instead of the widely used Poissonian-Gaussian noise model  (\emph{i.e.}, heteroscedastic Gaussian model~\citep{foi2008practical}).
It can accurately characterize the real noise structures by taking into account many noise sources, including photon shot noise, read noise, banding pattern noise, and quantization noise.

\iftrue
\begin{table*}[t]
\centering
\caption{Ablation study of different mechanisms for feature denoising during feature map downsampling. 
}
\scalebox{1.08}{
\begin{tabular}{c|c|ccc|ccc|c|c|cccc}
\hline
Filter type &Kernel size & AP& AP$_{50}$ & AP$_{75}$ & AP$^{box}$& AP$^{box}_{50}$ & AP$^{box}_{75}$ & Disturbance & GFlops & Parameters\\
\hline
None  & - &38.0 & 59.9 &39.1 &45.2 &67.1 &50.2 &1.5292 & 109.95 &\color{black} 43.78M  \\
\hline
Gaussian & $3\times 3$ &38.3 &60.5 &39.1 &45.5 &67.2&49.1 &1.4264 &109.96 &\color{black} 43.78M \\
Bilateral & $3\times 3$ &38.1 &59.3 &38.5 &45.3 &66.7&49.3 &1.5288 &109.96 &\color{black} 43.78M \\
Mean & $3\times 3$ &38.5 &60.4 &38.4 &45.7 &67.1&50.9 &1.4524 & 109.96 & \color{black} 43.78M \\
\hline
Spatial-variant & $3\times 3$ &39.0 & 61.0 &39.8 &46.3 &\color{black}\bf 68.0 &51.3 & 1.4011 &110.08 &\color{black} 43.93M \\
AWD & $3\times 3$ &\bf 39.3 &\bf 61.4 &\bf 40.2 &\bf 46.4 &\bf 68.0 &\bf 51.6 &\bf 1.3715 &110.25 &\color{black} 44.65M \\

\hline
\end{tabular}
}
\label{tab:downsample}
\end{table*}
\else
\begin{table*}[t]
\centering
\caption{Ablation study of different mechanisms for feature denoising. 
``AWD" and ``DSL" denote the proposed adaptive weighted downsampling layer and disturbance suppression learning{\color{black}, respectively}.
Disturbance is calculated by Eq~\eqref{eq:disturbance}.
}
\scalebox{1.2}{
\begin{tabular}{c|c|ccc|ccc|c|ccccc}
\hline
Filter type &Kernel size & AP& AP$_{50}$ & AP$_{75}$ & AP$^{box}$& AP$^{box}_{50}$ & AP$^{box}_{75}$ & GFlops & Parameters\\
\hline
None  & - &38.0 & 59.9 &39.1 &45.2 &67.1 &50.2 &109.95 & 43.78M  \\
\hline
Gaussian & $3\times 3$ &38.3 &60.5 &39.1 &45.5 &67.2&49.1  &109.96 &43.78M \\
Bilateral & $3\times 3$ &38.1 &59.3 &38.5 &45.3 &66.7&49.3  &109.96 &43.78M\\
Mean & $3\times 3$ &38.5 &60.4 &38.4 &45.7 &67.1&50.9 & 109.96 & 43.78M\\
\hline
Spatial-variant & $3\times 3$ &39.0 & 61.0 &39.8 &46.3 &68.0 &51.3 &110.08 &43.93M \\
AWD & $3\times 3$ &39.3 & 61.4 &40.2 &46.4 &68.0 &51.6  &110.25 &44.65M \\
AWD + DSL  & $3\times 3$ &\textbf{39.8} & \textbf{62.3} &\textbf{41.0} &\textbf{47.1} &\textbf{68.9} &\textbf{52.2}  &110.25 &44.65M\\

\hline
\end{tabular}
}
\label{tab:downsample}
\end{table*}
\fi

\subsection{Adaptive Weighted Downsampling Layer}
\label{sec:awd}
To be robust to image noise, the features of networks should be clean and consistently respond to the scene content.
As shown in Figure~\ref{fig:observation}{\color{red}(a)}, noise in low-light images introduces high-frequency disturbance in feature maps of convolutional neural networks, which can mislead the following semantic information extraction and degrade the final prediction.
We observe that the feature map downsampling is done by $1\times 1$ convolution layers with a stride of 2 in wide-used vanilla ResNet~\citep{resnet2016}, as shown in Figure~\ref{fig:resnet}{\color{red} (b)}.
This is similar to applying nearest neighbor interpolation for downsampling, which only considers the value of a single ``pixel".
It helps to reduce the computational cost but is useless for suppressing the noise in features. 
To better understand this, we show an example in Figure~\ref{fig:toy}.
We first obtain a noisy image from a clean one by injecting Gaussian noise ($\sigma=60$), then downsample the noisy image by nearest neighbor interpolation and mean filter{\color{black}, respectively}.
Due to the local smoothness prior, the mean filter is able to suppress image noise during downsampling, whereas nearest neighbor interpolation can do nothing with noise.
This applies to feature maps as well.

\vspace{+1mm}
\noindent\textbf{Downsample with low-pass filter.~}
On the basis of this analysis, 
we propose to use a low-pass filter (\eg, Gaussian filter, mean filter, or bilateral filter, with stride$=$2) for feature map downsampling.
\iftrue
To verify if it is helpful for feature noise suppression, we evaluate the feature noise with
\begin{equation}
\begin{aligned}
\label{eq:disturbance}
	D(x, x',f(\cdot;\theta)) = \sum_{i=1}^{n} 
	\lVert f^{(i)}(x;\theta) - f^{(i)}(x';\theta)\rVert^2_2,
\end{aligned}
\end{equation}
where $D$ indicates feature disturbance caused by image noise, \ie, feature noise, $x$ and $x'$ are clean normal-light image and corresponding noisy low-light image, and $f^{(i)}(x;\theta)$ is the $i$-th stage of feature maps in network $f(\cdot)$ with parameters of $\theta$.
\fi
As shown in Table~\ref{tab:downsample}, these low-pass filters are able to reduce the feature disturbance caused by noise in low-light images, and the instance segmentation performance also improves, which shows their effectiveness.

{\color{black}
\vspace{+1mm}
\noindent\textbf{Learning to generate spatial-variant filter.~}
Though these low-pass filters help to achieve better low-light instance segmentation results with minor extra computational cost, they are still suboptimal.
For example, they may blur edge/texture features with relatively high frequency in the scene content.
Furthermore, different spatial locations usually have different signal frequencies in feature maps. We need to apply different filters to them separately.
Therefore, we propose to use spatial-variant filters, which can be formulated as
\vspace{-1mm}
\begin{equation}
\label{eq:awd2}
\begin{aligned}
	Y_{i, j} &= \sum_{p,q\in S}
	W_{i, j}^{p,q} \cdot X_{i+p, j+q},
\end{aligned}
\end{equation}
where $X$, $Y$ are input and output feature maps, $(i, j)$ indicates the location in height, width dimensions, 
$S$ points to the set of spatial locations surrounding $(i, j)$, and $W$ is the filter weight predicted by the network
\begin{equation}
\color{black}
\vspace{-1mm}
\label{eq:awd}
\begin{aligned}
V_{i, j}\ &= \phi(X_{\Psi_{i, j}}), \\
W_{i, j}^{p,q}
&= \frac{\exp(V_{i, j}^{p,q})}{\sum_{p,q\in S} \exp(V_{i, j}^{p,q})},
\end{aligned}
\end{equation}
where $\phi$ is the weight generation function, $\Psi_{i, j}$  indexes the set of pixels $V_{i, j}$ conditioned on. 
The softmax function can ensure the filter kernels are low-pass.

\vspace{+1mm}
\noindent\textbf{Adaptive weighted downsampling layer.~}
Table~\ref{tab:downsample} shows the spatial-variant filter has superiority in feature denoising, but it still has two following drawbacks.
First, considering different channels of the feature maps show the semantic responses to different {\color{black}image features}, the signal frequencies can be variant across channels at the exact spatial locations.
Thus spatial-variant is not enough to achieve optimal results, it is necessary to generate spatial-variant and channel-variant filters for different spatial and channel locations.
Second, the spatial-variant filter generates filter weights from local features and fails to utilize the context and global information.
To solve these problems, we propose an adaptive weighted downsampling layer, which can be formulated as
\vspace{-1mm}
\begin{equation}
\label{eq:awd3}
\begin{aligned}
	Y_{c, i, j} &= \sum_{p,q\in S}
	W_{c, i, j}^{p,q} \cdot X_{c, i+p, j+q},
\end{aligned}
\end{equation}
where $(c, i, j)$ indicates the location in the channel, height, and width dimensions.
The $W$ is the filter weight predicted by the network
\begin{equation}
\color{black}
\vspace{-1mm}
\label{eq:awd4}
\begin{aligned}
V_{c, i, j}\ &= \phi(X_{\Psi_{c, i, j}}), \quad T_{c} = \phi'(\text{GP}(X)), \\
W_{c, i, j}^{p,q}
&= \frac{\exp(V_{c, i, j}^{p,q} \cdot T_{c})}{\sum_{p,q\in S} \exp(V_{c, i, j}^{p,q} \cdot T_{c})},
\end{aligned}
\end{equation}
{\color{black}
where $\phi$, $\phi'$ are also the weight generation functions, $\Psi_{c, i, j}$ indexes the set of pixels $V_{c, i, j}$ conditioned on, and $\text{GP}$ is the global pooling operation.
As illustrated in Figure~\ref{fig:resnet}{\color{red}(d)}, it is estimated by combining the local information $V$ (by using local features) and global information $T$ (by using global pooling features).
The predicted $T_{c}$ can adjust the smoothness of kernel $W_{c, i, j}$, which is similar to the temperature parameter in softmax~\citep{hinton2015distilling}.
To ensure that generated filters are low-pass, we use softmax to constrain weights to be positive and sum to 1.
In this way, we predict content-aware low-pass filters for each position and channel, so as to keep the foreground signal and suppress the feature noise adaptively.
} 

\begin{figure*}[t!]
\centering
\includegraphics[width=0.95\linewidth]{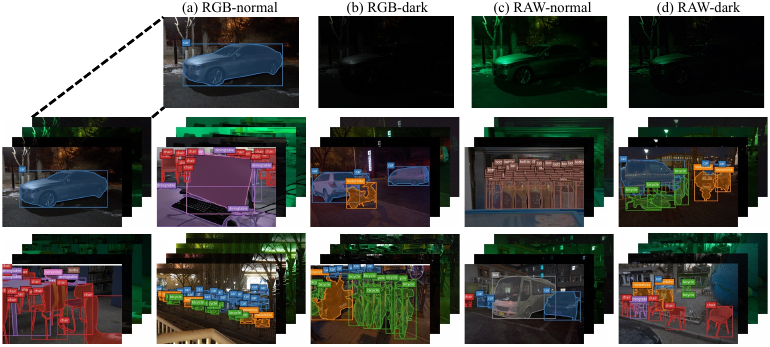}
\caption{\small Example scenes in our LIS dataset.  
Four image types (long-exposure normal-light and short-exposure low-light images in both RAW and sRGB formats) are captured for each scene.}
\label{fig:LIS_samples}
\end{figure*}

\subsection{Smooth-Oriented Convolutional Block}
\label{sec:scb}
The adaptive weighted downsampling layer improves the robustness of networks with a carefully designed downsampling process.
To further enhance the networks, we turn to focus on increasing the strength of ordinary convolutional blocks.
The principles of convolutional block design are two sides.
First, the improved convolutional block should be more robust to the feature noise.
Second, the extra computational cost should be minor or free.
To this end, we propose the smooth-oriented convolutional block.
It can replace the ordinary $3\times3$ convolutional layers to enhance the network.
Remarkably, it brings no extra computational cost during inference by using the re-parameterization technique~\citep{ding2021repvgg}.

The structure of the smooth-oriented convolutional block is shown in Figure~\ref{fig:scb}, it explicitly employs a linear combination of multiple branches during the training stage. 
The main convolutional branch is the same as the original $3\times3$ convolutional layers.
The auxiliary branch consists of a $1\times1$ convolutional layer and a smooth-oriented convolution, which can learn smooth kernels to suppress the feature noise in high-frequency.
To ensure the filters are smooth, we use the mean filter or Gaussian filter to initialize the weights of smooth-oriented convolution and regularize its weights with the softmax function, so as to ensure that the learned $3\times 3$ kernel for each channel is all positive and sum to 1. 
And it is followed by a $1\times1$ convolutional layer to fuse the filtered feature into the main convolutional branch.

And in the inference stage, the smooth-oriented convolutional block can be folded back to a normal $3\times3$ convolutional layer by the re-parameterization technique~\citep{ding2021repvgg}.
Formally, we use $W\in R^{C_{2}\times C_{1}\times3\times3}$ to denote the kernel of a $3 \times 3$ convolutional layer with $C_1$ input channels and $C_2$ output channels.
The kernel of folded $3\times3$ convolutional can be re-parameterized as follows
\begin{equation}
\begin{aligned}
\vspace{-2mm} 
	W'_{3\times 3}[i, j, h, t] & = W_{3\times 3}[i, j, h, t] \\
	& + (W_{1\times 1}[i, j, 1, 1] * W_{\text{SConv}}[i, 1, h, t]),
\vspace{-2mm}
\end{aligned}
\end{equation}
where $i\in\{1, 2, ...,C_2\}$, $j\in\{1, 2, ...,C_1\}$, and $h, t \in\{1, 2, 3\}$ are indicators.
$W'_{3\times 3} \in R^{C_2\times C_1\times 3\times 3}$ is the kernel weights of folded convolutional layer for inference,
and $W_{3\times 3} \in R^{C_2\times C_1\times 3\times 3}$,
$W_{1\times 1} \in R^{C_2\times C_1\times 1\times 1}$ 
and $W_{\text{SConv}} \in R^{C_2\times 1\times 3\times 3}$ indicate $3\times 3$ convolutional layer, $1\times 1$ convolutional layer and smooth-oriented convolution during training, respectively.

\begin{table*}[t]
\footnotesize
\caption{
Ablation study for the low-light RAW synthetic pipeline on our LIS testing set.
``UP" and ``NI" indicate unprocessing and noise injection operations{\color{black}, respectively}. 
The ``*" indicates processing with gamma correction.
All models are based on the vanilla Mask R-CNN~\citep{MaskRCNN2017} with the backbone of ResNet-50-FPN~\citep{resnet2016,2017feature}.
}
\label{tab:abl-syn}
\centering
\newcommand{\tabincell}[2]{\begin{tabular}{@{}#1@{}}#2\end{tabular}}

\scalebox{0.92}{
\begin{tabular}{c|c|c|cc|ccc|ccccc}
\hline
\multirow{2}{*}{\tabincell{c}{Data Type}} & \multirow{2}{*}{\tabincell{c}{Training Set}} & \multirow{2}{*}{\tabincell{c}{Testing Set}}  
& \multirow{2}{*}{UN} & \multirow{2}{*}{NI} 
& \multirow{2}{*}{AP} & \multirow{2}{*}{AP$_{50}$ } & \multirow{2}{*}{AP$_{75}$ } & \multirow{2}{*}{AP$^{box}$} & \multirow{2}{*}{AP$^{box}_{50}$ } & \multirow{2}{*}{AP$^{box}_{75}$ } \\
&  & &  & & & &  &  & \\
\hline
\multirow{5}{*}{REAL}
& LIS sRGB-normal   & LIS sRGB-normal &-&-&48.1 &71.8 &50.3 &54.6&76.5&60.2\\
& LIS RAW-normal & LIS RAW-normal &-&-&45.4 &70.0 &46.6 &52.8 &74.7 &63.4 \\
&\color{black}  LIS RAW-normal* &\color{black}  LIS RAW-normal* &-&- &48.1 &71.5 &50.5 &54.7&76.0&60.3 \\
\cline{2-11}
& LIS sRGB-dark   & LIS sRGB-dark 
& -  & -&35.5 &57.5 &36.1  & 42.9 & 64.3 & 46.1  \\
& LIS RAW-dark   & LIS RAW-dark 
& -  & - &39.0 &61.3 &40.1 &46.1 &67.8 &50.5 \\
\cline{1-11}
\multirow{8}{*}{SYNTHETIC}
& \multirow{4}{*}{COCO sRGB-normal}
& \multirow{4}{*}{LIS RAW-dark}    
& -  & -  &23.2   &40.0  &22.5 &26.1  &42.7  &27.3  \\
& & & $\surd$  & - &23.4 &38.0 &23.3 &26.6 &42.5 &28.7 \\
& & & -  & $\surd$ &27.2 &46.3 &27.7  &31.0 &50.7 &32.6 \\
& &  & $\surd$  & $\surd$  &\bf 29.4 &\bf 49.0 &\bf 28.5 &\bf 34.4 &\bf 54.6 &\bf 36.6 \\ 
\cline{2-11}
& \multirow{4}{*}{LIS sRGB-normal} 
& \multirow{4}{*}{LIS RAW-dark}
& - & - &31.6 &50.2 &31.6 &36.6 &56.0 &39.3  \\
& & & $\surd$ & - &33.8 &52.7 &34.2 &39.6 &58.8 &43.2 \\
& & & -  & $\surd$ &35.1 &55.5 &35.4 &40.6 &61.1 &44.1 \\
& &  & $\surd$  & $\surd$ &\bf 38.0 &\bf 59.9 &\bf 39.1 &\bf 45.2 &\bf 67.1 &\bf 50.2 \\ 
\hline

\end{tabular}}
\vspace{-3mm}
\end{table*}

\subsection{Disturbance Suppression Learning}
\label{sec:dsl}
Ideally, a robust network should extract similar features regardless of whether the input image is corrupted by noise or not. 
Orthogonal to the architectural considerations, we introduce disturbance suppression learning to encourage the network to learn disturbance-invariant features during training. 
As shown in Figure~\ref{fig:overview}, the total loss for learning is 
\begin{equation}
\begin{aligned}
\color{black}
	L(\theta) = L_{\text{IS}}(x;\theta) 
	+ \alpha L_{\text{IS}}(x';\theta) + \beta L_{\text{DS}}(x, x';\theta),
\end{aligned}
\end{equation}
where $x$ is the unprocessed clean synthetical RAW image and $x'$ is its noisy version,
$\alpha$ and $\beta$ are the weights of the losses. 
We empirically set $\alpha$, $\beta$ to 1, 0.01 for weighing.
The $L_{\text{IS}}$ is instance segmentation loss, which consists of classification loss, bounding box regression loss, and segmentation (per-pixel classification) loss. 
Its specific formula is related to the instance segmentation model, please refer to Mask R-CNN~\citep{MaskRCNN2017} for details.
The model should learn to work stably whether the image is noisy or not. 
Hence $L_{\text{IS}}$ is applied to both clean image $x$ and noisy image $x'$.
The $L_{\text{DS}}$ is feature disturbance suppression loss, which is defined as
\begin{equation}
\begin{aligned}
\vspace{-2mm} 
\color{black}
	L_{\text{DS}}(x, x';\theta) = \sum_{i=1}^{n} 
	\lVert f^{(i)}(x;\theta) - f^{(i)}(x';\theta)\rVert^2_2,
\vspace{-2mm}
\end{aligned}
\end{equation}
where $f^{(i)}(x;\theta)$ is the $i$-th stage of feature maps of model.
By minimizing the Euclidean distance between clean features $f^{(i)}(x;\theta)$ and noisy features $f^{(i)}(x';\theta)$, disturbance suppression loss induces model to learn disturbance-invariant features. 
Therefore the feature disturbance caused by image noise can be reduced, and its robustness for corrupted low-light images is improved.

Different from perceptual loss~\citep{2020qis}, we do not need to pretrain a teacher model, which makes our training simpler and faster.
With $L_{\text{IS}}(x;\theta)$, $L_{\text{IS}}(x';\theta)$, our model can learn discriminative features from both clean and noisy images, so as to keep stable accuracy no matter images are corrupted by noise or not.
Whereas ``student" in perceptual loss~\citep{2020qis} only sees noisy images, which leads to degradation on clean images and limits its robustness. 
Moreover, the domain gap of feature distribution between the teacher model and student model may harm the learning procedure. 
While we minimize the distance between clean features and noisy features predicted by the same model, which avoids this problem.

 \begin{table*}[t!]
 \centering
 \caption{
 Ablation study for adaptive weighted downsampling (AWD), smooth-oriented convolutional block (SCB), and disturbance suppression learning (DSL).
``Synthetic LIS" indicates using unprocessing and noise injection operations to synthetic low-light images from sRGB-normal images for training.
 ``Real LIS" indicates the RAW-dark and RAW-normal image pairs in the training set are accessible for training.
 Results are reported on the LIS RAW-dark test set.
 }
 \vspace{-2mm}
 \scalebox{.92}{
 \begin{tabular}{c|c|ccc|ccc|ccc|c|cccc}
 \hline
\multirow{2}{*}{Training Data}& \multirow{2}{*}{Backbone} & \multirow{2}{*}{AWD} & \multirow{2}{*}{SCB} & \multirow{2}{*}{DSL} 
& \multirow{2}{*}{AP} & \multirow{2}{*}{AP$_{50}$ } & \multirow{2}{*}{AP$_{75}$ } & \multirow{2}{*}{AP$^{box}$} & \multirow{2}{*}{AP$^{box}_{50}$ } & \multirow{2}{*}{AP$^{box}_{75}$ } & \multirow{2}{*}{GFlops} & \multirow{2}{*}{Parameters}\\
&  & &  & & & &  &  & &  & \\
 \hline

 \multirow{5}{*}{Synthetic LIS}
 & ResNet-50-FPN & - & - & -  & 38.0 & 59.9 &39.1 &45.2 &67.1 &50.2 &109.95 &\color{black}43.78M \\
 & ResNet-101-FPN & - & - & -  & 39.5 & 62.2 &40.5 &46.7 &68.4 &52.7 &128.01 &\color{black}62.78M \\
 \cline{2-13}
 & ResNet-50-FPN & $\surd$ & - & - &39.3 &61.4 &40.2 &46.4 &68.0 &51.6 &110.25 &\color{black}44.65M\\
 & ResNet-50-FPN & $\surd$ & $\surd$ & - &39.9 &61.8 &41.1 &47.3 &68.8 &52.3 &110.25 &\color{black}44.65M\\
 & ResNet-50-FPN & $\surd$ & $\surd$ & $\surd$ &\bf 40.8 &\bf 62.7 &\bf 41.5 &\bf 48.0 &\bf 69.2 &\bf 52.6 &110.25 &\color{black}44.65M\\
 \hline 
 \multirow{5}{*}{Real LIS} 
 & ResNet-50-FPN & - & - & -  &39.0 & 61.3 &40.1 &46.1 &67.8 &50.5 &109.95 & \color{black}43.78M \\
 & ResNet-101-FPN & - & - & -  &40.7 & 63.6 &41.4 &48.7 &70.2 &53.5 &128.01 &\color{black}62.78M \\
 \cline{2-13}
 & ResNet-50-FPN & $\surd$ & - & - & 41.0 &63.6 &41.5 &48.6 &70.8 &51.9  &110.25 &\color{black}44.65M \\
 & ResNet-50-FPN & $\surd$ & $\surd$ & - &41.5 &64.3 &41.9 &48.9 &71.6 &52.7 &110.25 &\color{black}44.65M\\
 & ResNet-50-FPN & $\surd$ & $\surd$ & $\surd$ &\bf 42.7 &\bf 66.2 &\bf 43.3 &\bf 50.3 &\bf 72.6 &\bf 55.2 &110.25 &\color{black}44.65M\\
 \hline
 \end{tabular}
 }
 \vspace{-3mm}
 \label{tab:AWD_SCB_DSL}
 \end{table*}

\section{Low-light Instance Segmentation Dataset}
Though evaluating on synthetic low-light images is a common and convenient practice~\citep{2021meat}, its results can severely deviate from the real world due to the much more complicated lighting conditions and image noise~\citep{2017benchmarking, 2018renoir}.
To reveal and systematically investigate the effectiveness of the proposed method in the real world, a real low-light image dataset for instance segmentation is necessary and urgently needed.
Considering there is no suitable dataset, therefore, we collect and annotate a Low-light Instance Segmentation (\textbf{LIS}) dataset using a Canon EOS 5D Mark IV camera.
In Figure~\ref{fig:LIS_samples}, we show some examples of annotated images in our LIS dataset.
It exhibits the following characteristics:
\begin{itemize}
\item \textbf{Paired samples.~} In the LIS dataset, we provide images in both sRGB-JPEG (typical camera output) and RAW formats, each format consists of paired short-exposure low-light and corresponding long-exposure normal-light images. We term these four types of images as \textit{sRGB-dark, sRGB-normal, RAW-dark, and RAW-normal}. To ensure they are pixel-wise aligned, we mount the camera on a sturdy tripod and avoid vibrations by remote control via a mobile app.  
\item \textbf{Diverse scenes.~} The LIS dataset consists of 2230 image pairs, which are collected in various scenes, including indoor and outdoor. To increase the diversity of low-light conditions, we use a series of ISO levels (\eg, 800, 1600, 3200, 6400) to take long-exposure reference images, and we deliberately decrease the exposure time by a series of low-light factors (\eg, 10, 20, 30, 40, 50, 100) to take short-exposure images for simulating very low-light conditions. 
\item \textbf{Instance-level pixel-wise labels.~} For each pair of images, we provide precise instance-level pixel-wise labels annotated by professional annotators, yielding 10504 labeled instances of 8 most common object classes in our daily life (bicycle, car, motorcycle, bus, bottle, chair, dining table, tv).
\end{itemize}

We note that LIS contains images captured in different scenes (indoor and outdoor), and different illumination conditions.
In Figure~\ref{fig:LIS_samples}, object occlusion and densely distributed objects make LIS more challenging besides the low light.

\section{Experiments}
In this section, we first introduce implementation details and evaluation metrics.
Then we conduct ablation studies to evaluate the effectiveness of the proposed method.
Finally, we compare our method against existing multi-step methods.

\subsection{Implementation Details}

All experiments here are conducted on {\color{black}Mask R-CNN~\citep{MaskRCNN2017}} baseline with ResNet-50-FPN~\citep{resnet2016, 2017feature} backbone for simplicity.  
Notice that our proposed method can be equipped with any network-based instance segmentation model. 

\vspace{+1mm}
\noindent\textbf{Training details.~}
Our framework is trained by synthetic low-light RAW-RGB\footnote{
To make the detector compatible with sRGB inputs, instead of the Bayer RAW images, we follow~\citep{chen2019seeing} to use demosaicked 3-channel RAW-RGB images as inputs, where the green channel is obtained by averaging the two green pixels in each two-by-two Bayer block. 
In the following, {\color{black}we refer to} "RAW" and "RAW-RGB" interchangeably.} images generated from COCO~\citep{2014microsoft} dataset using our low-light RAW synthetic pipeline\footnote{We use COCO samples belonging to the same 8 object classes in the LIS dataset.}. 
Our implementation is based on \textit{MMDetection}.
During training, we use random flip as data augmentation and train with a batch size of 8, a learning rate of 1e-2 for 12 epochs, with a learning rate dropping by 10$\times$ at 8 and 11 epochs, respectively.  
To make the model quickly adapt to low-light settings, we use COCO pre-trained model as initialization.

\begin{table}[t!]
\centering
\caption{
\small
Ablation study for various bit-depth and encodings.
The {\it sRGB-dark} with italics indicates images are obtained from corresponding RAW images.
It is worth noting that we simulate various color encodings by quantizing the captured 14-bit RAW images to RAW images of different color encodings (\eg, 8, 10, and 12 bits).
This can be different from directly capturing images in corresponding bits.
}
\scalebox{0.8}{
\begin{tabular}{l|ccc|cccc}
\hline
Data type & AP& AP$_{50}$ & AP$_{75}$ & AP$^{box}$& AP$^{box}_{50}$ & AP$^{box}_{75}$  \\
\hline
sRGB-dark (8-bit) &35.5 &57.5 &36.1  & 42.9 & 64.3 & 46.1 \\
\it sRGB-dark (10-bit) & 37.7 & 59.5 & 38.3 & 44.5 & 66.5 & 48.7 \\
\it sRGB-dark (12-bit) & 38.2 & 60.4 & 39.3 & 45.3 & 67.2 & 49.0 \\
\it sRGB-dark (14-bit) & 38.7 & 60.8 & 39.5 & 46.0 & 67.4 & 50.8 \\
\hline
RAW-dark (8-bit)  & 35.5 & 58.2 & 35.8 & 42.7 & 65.6 & 45.2 \\
RAW-dark (10-bit) & 38.6 & 60.7 & 39.9 & 45.9 & 67.6 & 50.0 \\
RAW-dark (12-bit) & 39.0 &61.4 &40.0 &46.0 &68.5 &50.3\\
RAW-dark (14-bit) &39.0 &61.3 &40.1 &46.1 &67.8 &50.5 \\
\hline
\end{tabular}
}
\label{tab:bits}
\end{table}

\begin{table}[t!]
\color{black}
\centering
\caption{
\small
Ablation for global pooling branch in adaptive weighted downsampling layer.
}
\scalebox{0.8}{
\begin{tabular}{l|ccc|cccc}
\hline
Method & AP& AP$_{50}$ & AP$_{75}$ & AP$^{box}$& AP$^{box}_{50}$ & AP$^{box}_{75}$  \\
\hline
w/ Global Pooling  &39.1 & 61.1 &39.8 &46.3 &\bf 68.0 &51.4 \\
w/o Global Pooling & \bf 39.3 &\bf 61.4 &\bf 40.2 &\bf 46.4 &\bf 68.0 &\bf 51.6 \\
\hline
\end{tabular}
}
\label{tab:gp}
\end{table}

\begin{table}[t!]
\centering
\caption{
Ablation study for kernel size of adaptive weighted downsampling layer. 
We use synthetic LIS training set for training.
Results are reported on the LIS test set.
}
 \scalebox{0.9}{
 \begin{tabular}{c|ccc|cccc}
 \hline
 Kernel size & AP& AP$_{50}$ & AP$_{75}$ & AP$^{box}$& AP$^{box}_{50}$ & AP$^{box}_{75}$  \\
 \hline
 None &38.0 &59.9 &39.1 &45.2 &67.1 &50.2 \\
 \hline
 $2\times 2$ &38.9 &61.3 &39.1 &\bf 46.4 &\bf 68.1 &51.0 \\
 $3\times 3$ &\bf 39.3 & \bf 61.4 & \bf 40.2 &\bf 46.4 &68.0 &\bf 51.6 \\
 $4\times 4$ &38.9 &\bf 61.4 &39.5 &\bf 46.4 &\bf 68.1 &50.2 \\
 $5\times 5$ &38.4 &60.5 &39.4 &45.5 &67.6 &50.6 \\
 \hline
 \end{tabular}
 }
 \label{tab:kernel}
 \end{table}

\begin{table}[t!]
\color{black}
\centering
\caption{
\small
Comparison and combination with different attention mechanisms.
The baseline model is Mask R-CNN~\citep{MaskRCNN2017} with ResNet-50-FPN~\citep{resnet2016,2017feature}. 
We compare and combine the proposed Adaptive Downsampling (AWD) layer with typical spatial attention (\ie, non-local~\citep{2018nonlocal}), channel attention (\ie, squeeze-and-excitation (SE) ~\citep{2018senet}) and both spatial and channel attention (\ie, CBAM~\citep{2018cbam}).
Models are trained on synthetic LIS and evaluated on the LIS test set.
$\Delta$$\uparrow$ indicates the extra improvement brought by the AWD. 
\vspace{+1mm}
}
\scalebox{0.8}{
\begin{tabular}{l|ccc|ccc}
\hline
Method & AP& AP$_{50}$ & AP$_{75}$ & AP$^{box}$& AP$^{box}_{50}$ & AP$^{box}_{75}$  \\
\hline
baseline & 38.0 & 59.9 & 39.1 & 45.2 & 67.1 & 50.2 \\ 
+ AWD &39.3 &61.4 &40.2 &46.4 &68.0 &51.6 \\ 
$\Delta$$\uparrow$ &\bf +1.3 &\bf +1.5 &\bf +1.1 &\bf +1.2 &\bf +0.9 &\bf +1.4 \\
\hline
Non-local &38.5 &60.0 &39.8 &45.6 &67.4 & 50.5 \\
Non-local + AWD &39.5 &61.4 &40.2 &46.6 &68.8 & 51.7 \\
$\Delta$$\uparrow$ &\bf +1.0 &\bf +1.4 &\bf +0.4 &\bf +1.0 &\bf +1.4 &\bf +1.2 \\
\hline
SE &39.2 &61.4 &40.3 &46.1 &67.8 & 52.0 \\
SE + AWD &40.3 &62.8 &41.4 &47.2 &69.4 & 52.5 \\
$\Delta$$\uparrow$ &\bf +1.1 &\bf +1.4 &\bf +1.1 &\bf +1.1 &\bf +1.6 &\bf +0.5 \\
\hline
CBAM &38.7 &60.7 &40.0 &45.7 &67.3 & 51.1 \\
CBAM + AWD &40.0 &62.8 &40.5 &47.4 &69.6 & 52.8 \\
$\Delta$$\uparrow$ &\bf +1.3 &\bf +2.1 &\bf +0.5 &\bf +1.7 &\bf +2.3 &\bf +1.7 \\
\hline
\end{tabular}
}
\label{tab:attention}
\end{table}

\begin{figure}[t!]
\centering
\begin{tabular}{ccccc}
\hspace{-4mm}
\includegraphics[width=0.31\linewidth]{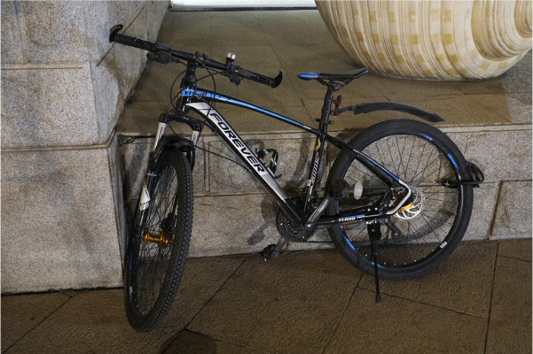}\hspace{-4mm}
&\includegraphics[width=0.31\linewidth]{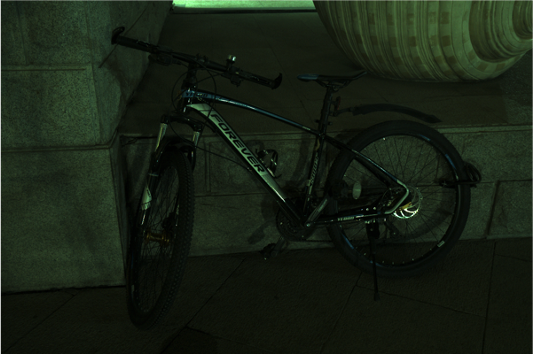}\hspace{-4mm}
&\includegraphics[width=0.31\linewidth]{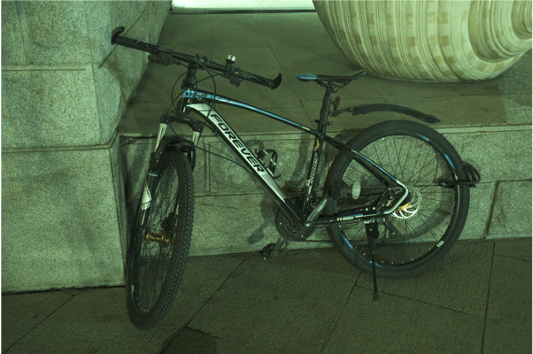} \\
(a) \hspace{-4mm} & (b) \hspace{-4mm} & (c) \\
\end{tabular}
\caption{
\small
Illustration of sRGB-normal and RAW-normal with or without gamma correction.
The sRGB-normal (a) is much visually brighter than the RAW-normal (b), especially for the dark region.
And after gamma correction, RAW-dark (c) shows similar illumination to sRGB-normal.
}
\label{fig:gamma}
\end{figure}

\begin{figure*}[t!]
\centering
\scalebox{0.92}{
\begin{tabular}{cccccccccc}
\includegraphics[width=0.20\linewidth]{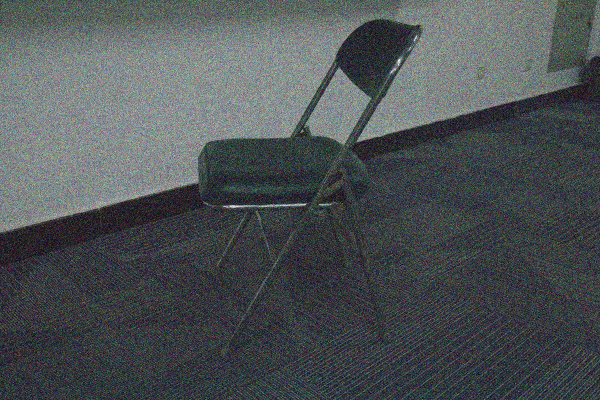}\hspace{-3mm}
&\includegraphics[width=0.20\linewidth]{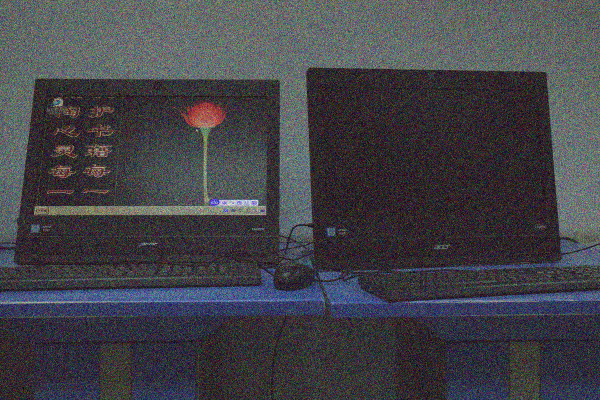}\hspace{-3mm}
&\includegraphics[width=0.20\linewidth]{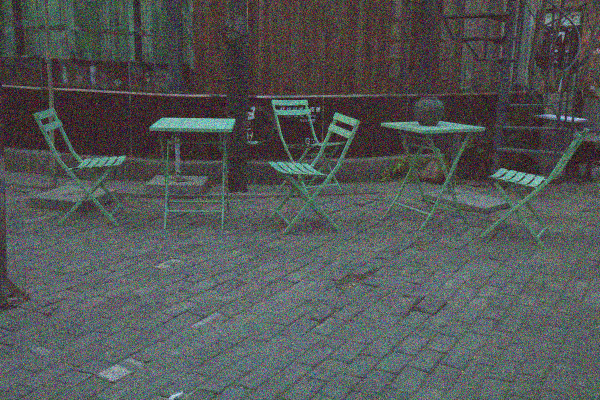}\hspace{-3mm}
&\includegraphics[width=0.20\linewidth]{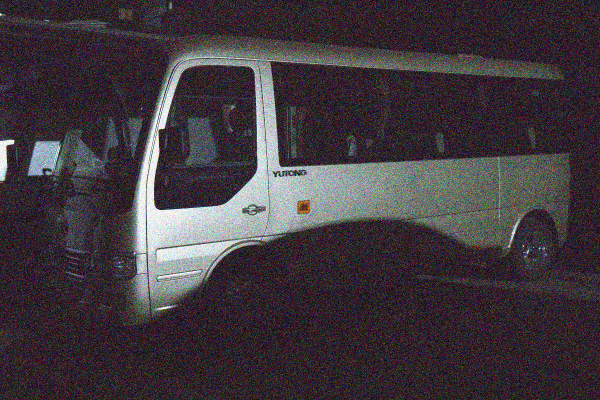}\hspace{-3mm}
&\includegraphics[width=0.20\linewidth]{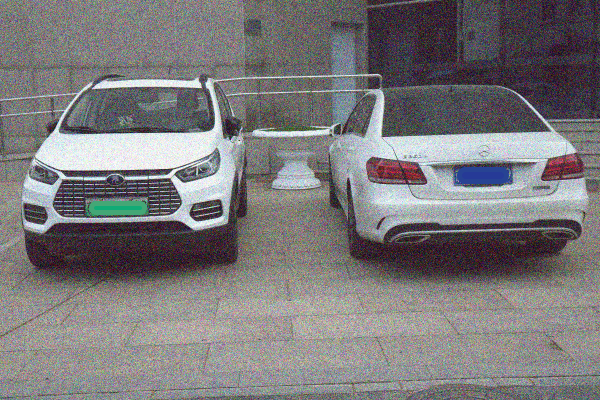} 
\\
\includegraphics[width=0.20\linewidth]{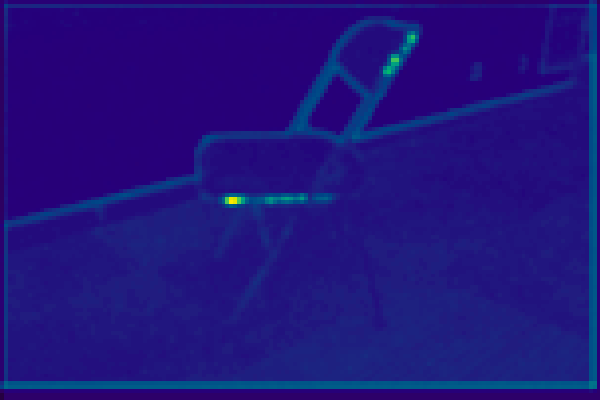}\hspace{-3mm}
&\includegraphics[width=0.20\linewidth]{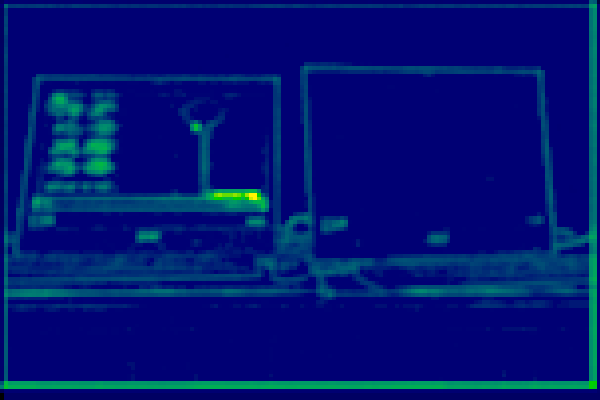}\hspace{-3mm}
&\includegraphics[width=0.20\linewidth]{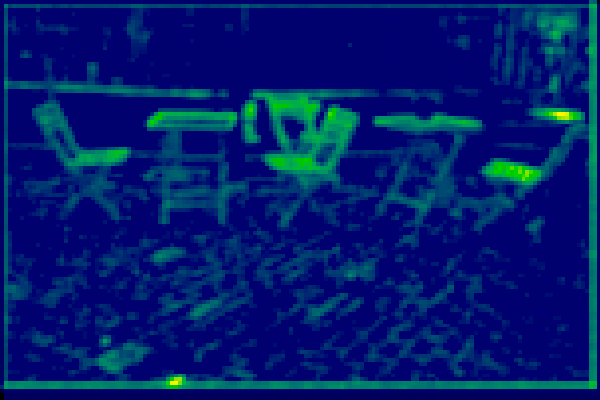}\hspace{-3mm}
&\includegraphics[width=0.20\linewidth]{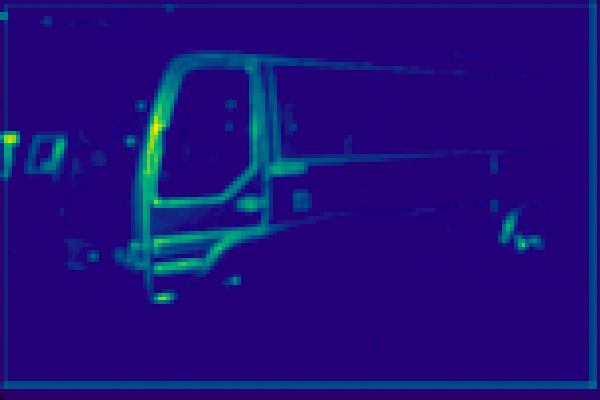}\hspace{-3mm}
&\includegraphics[width=0.20\linewidth]{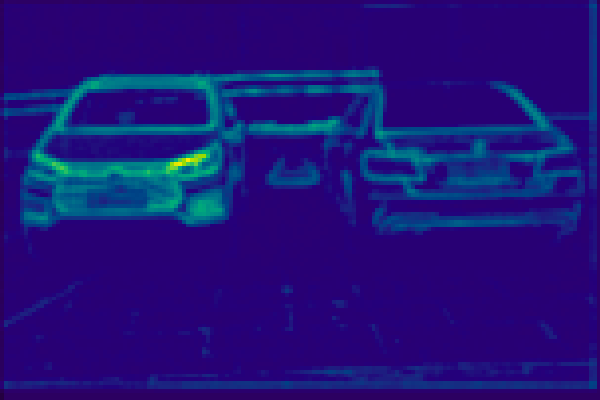}
\\

\end{tabular}
}
\caption{
Visualization of filter weights predicted by adaptively weighted downsampling (AWD) layer.
The top row shows input images, we convert RAW images to sRGB images for visualization.
The bottom row shows the visualized standard variance of learned filter weights for each position, brighter color means a higher standard variance of predicted filter weights.
}
\label{fig:learned_kernel}
\end{figure*}


 \begin{table}[t!]
 \centering
 \caption{
 Ablation study for the smooth-oriented convolutional block.
 SConv indicates smooth-oriented convolution.
 We use synthetic LIS training set for training.
 Results are reported on the LIS test set.
 }
 \scalebox{0.82}{
 \begin{tabular}{c|ccc|cccc}
 \hline
 Method & AP& AP$_{50}$ & AP$_{75}$ & AP$^{box}$& AP$^{box}_{50}$ & AP$^{box}_{75}$  \\
 \hline
Baseline (/w AWD) &39.3 &61.4 &40.2 &46.4 &68.0 &51.6 \\
 \hline
Gaussian &39.4 &61.6 &39.8   &46.5 &68.2 &51.4 \\
Mean &39.4 &61.7 &40.2   &46.6 &68.6 &51.4 \\
SConv &\textbf{39.9} &\textbf{61.8} &\textbf{41.1} &\textbf{47.3} &\textbf{68.8} &\textbf{52.3} \\ 
 \hline
 \end{tabular}
 }
 \label{tab:scb}
 \end{table}

 \begin{table}[t!]
 \color{black}
 \centering
 \caption{
 Ablation study for disturbance suppression learning (DSL) and perceptual loss (PL). 
 We use synthetic LIS training set for training.
 Results are reported on the LIS test set.
 }
 \scalebox{0.96}{
 \begin{tabular}{c|ccc|cccc}
 \hline
 \multirow{2}{*}{Method} & \multicolumn{3}{c|}{RAW-dark} & \multicolumn{3}{c}{RAW-normal} \\
 \cline{2-7}
  & AP& AP$_{50}$ & AP$_{75}$  & AP& AP$_{50}$ & AP$_{75}$  \\
 \hline
 Baseline &39.9 &61.8 &41.1 &44.7 &68.2 &45.7 \\
 \hline
 w/ PL &40.2 &61.9 &40.7   &45.1 &68.4 &45.9 \\
 w/ DSL &\textbf{40.8} &\textbf{62.7} &\textbf{41.5} &\textbf{46.6} &\textbf{69.5} &\textbf{48.6} \\ 
 \hline
 \end{tabular}
 }
 \label{tab:TSL}
 \end{table}

\begin{table}[t!]
\centering
\caption{
\small
Ablation study for adaptive weighted downsampling (AWD), smooth-oriented convolution block (SCB), and disturbance suppression learning (DSL) under normal-light conditions.
Models are trained on the COCO train set, and results are reported on the COCO val set.
}
\scalebox{0.80}{
\begin{tabular}{l|ccc|ccccc}
\hline
Method & AP& AP$_{50}$ & AP$_{75}$ & AP$^{box}$& AP$^{box}_{50}$ & AP$^{box}_{75}$ \\
\hline
Mask R-CNN 
&34.4 &55.6 &36.9 &38.0 &58.6 &41.5\\
\hline
+AWD &35.5 &56.6 &37.9 & 38.8& 59.7 & 42.5\\
+AWD+SCB &35.5 &56.6 &38.0 & 39.0& 59.9 & 42.5\\
+AWD+SCB+DSL &\bf 36.1 &\bf57.4 &\bf39.0 &\bf 39.5&\bf 60.3 &\bf 43.2\\
\hline
\end{tabular}
}
\label{tab:coco}
\end{table}

\begin{figure}[t!]
\centering
\scalebox{0.64}{
\begin{tabular}{cccccc}
\includegraphics[width=0.48\linewidth]{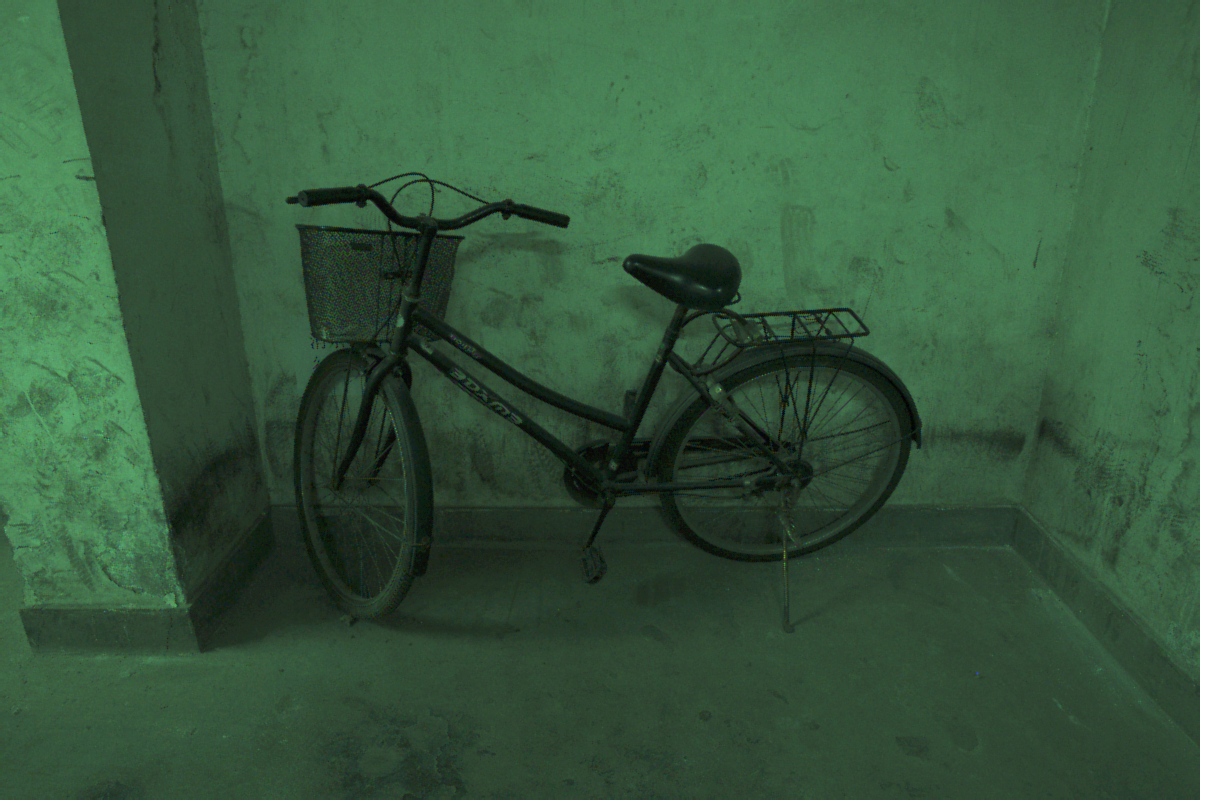}
\hspace{-3mm}
&\includegraphics[width=0.48\linewidth]{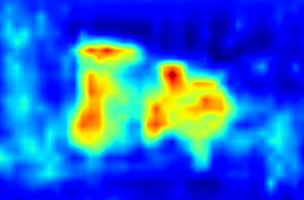}
\hspace{-3mm}
&\includegraphics[width=0.48\linewidth]{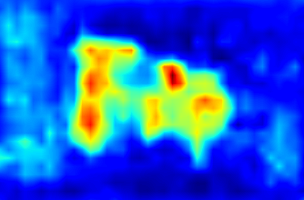}
\\
Original image & Clean feature (baseline) &Clean feature (proposed)
\\
\includegraphics[width=0.48\linewidth]{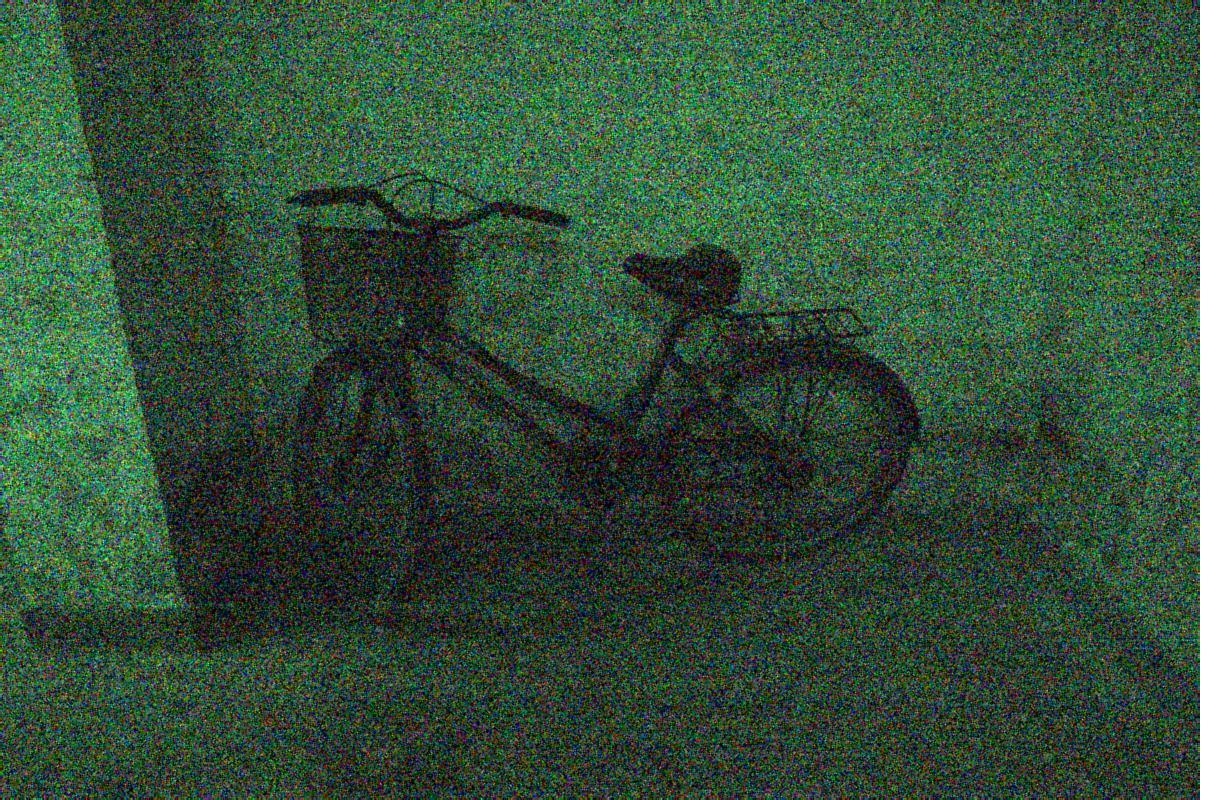}
\hspace{-3mm}
&\includegraphics[width=0.48\linewidth]{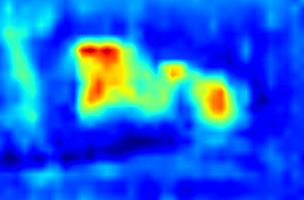}
\hspace{-3mm}
&\includegraphics[width=0.48\linewidth]{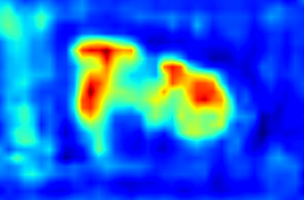}

\\
Noisy image & Noisy feature (baseline) & Noisy feature (proposed) 
\\
\includegraphics[width=0.48\linewidth]{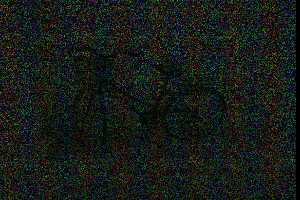}
\hspace{-3mm}
&\includegraphics[width=0.48\linewidth]{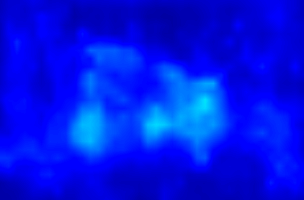}
\hspace{-3mm}
&\includegraphics[width=0.48\linewidth]{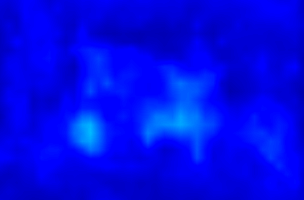}
\\
Image noise & Residual error (baseline) & Residual error (proposed)
\end{tabular}
}
\caption{\small 
Visualized high-level features.
The residual error shows the difference between the clean feature and the noisy feature.
It can be seen that the proposed method helps to reduce feature disturbance and keeps semantic responses to scene content when the image is noisy.
}
\label{fig:featuremap}
\end{figure}

\vspace{+1mm}
\noindent\textbf{Dataset and Evaluation Metrics.~}
The real low-light instance segmentation performance is evaluated on the LIS dataset, in which the total 2230 image pairs are randomly split into a train set of 1561 pairs and a test set of 669 pairs.

Following~\citep{MaskRCNN2017}, we measure the performance by using COCO-style AP (averaged over thresholds from 0.5 to 0.95 with an interval of 0.05), AP$_{50}$ and AP$_{75}$ {\color{black}(\ie, AP at an IoU of 0.5)}.
We also provide the results of detection, which are represented as AP$^{box}$, AP$^{box}_{50}$, and AP$^{box}_{75}$.
To evaluate inference speed, we {\color{black}measure} Frame Per Second (FPS) for each method on $600\times 400$ images with a single RTX 3090. 

\subsection{Ablation Studies}
{\color{black}
In this section, we first conduct ablation studies on input image types to reveal the advantage of RAW images for low-light instance segmentation and verify the effectiveness of low-light synthetic pipeline in Table~\ref{tab:abl-syn}.
}
Then, we investigate the adaptive weighted downsampling layer, smooth-oriented convolutional block, and disturbance suppression learning.
As shown in Table~\ref{tab:AWD_SCB_DSL}, they all contribute to performance improvement.
Finally, we verify the effectiveness of the low-light synthetic RAW pipeline.
All results are reported on the LIS test set.

\vspace{+1mm}
\noindent\textbf{sRGB \emph{vs.} RAW.~}
To explore the upper bound of sRGB image and RAW image under normal-light and low-light conditions, we experiment with sRGB-normal, RAW-normal, sRGB-dark, and RAW-dark separately on our LIS dataset.
As shown in Table~\ref{tab:abl-syn}, under normal light, sRGB and RAW have similar performance upper bound.
But there is a consistent performance gap between RAW-normal and sRGB-normal (45.4 AP \emph{vs.} 48.1 AP).
To investigate deeper into this phenomenon, 
we notice that though they contain similar scene information, the sRGB-normal is much visually brighter than the RAW-normal, especially for the dark region (see Figure~\ref{fig:gamma}).
This is caused by gamma correction in the pipeline of processing RAW to sRGB.
Owing to the non-linear manner in perception, the sensitivity of humans to relative differences between darker tones is more significant than between lighter tones.
And gamma correction can avoid allocating too many bits to highlights that humans cannot differentiate.
Generally, the gamma correction can be written as:
\vspace{-1mm}
\begin{equation}
\begin{aligned}
I_{out} = I_{in}^{\gamma}
\end{aligned}
\end{equation}
where $I\in [0, 1]$ is the normalized image pixel, and $\gamma$ is usually set as $1/2.2$ for processing RAW to sRGB~\citep{brooks2019unprocessing}.
After applying gamma correction, RAW-dark with gamma correction shows similar illumination as shown in Figure~\ref{fig:gamma} (c).
And the corresponding performance is consistent with sRGB-normal, as shown in Table~\ref{tab:abl-syn}.

As for low-light conditions, the result of RAW images largely outperforms that of sRGB images, showing that RAW images keep richer scene information under very low light.
This is critical for the instance segmentation task.
We assume it is due to the higher color encodings of RAW images (14-bit RAW from Canon EOS 5D \emph{vs.} 8-bit sRGB ).
To verify this, we simulate various color encodings by quantizing the captured 14-bit RAW images to RAW images of different color encodings (\eg, 8, 10, and 12 bits).
And the sRGB-dark of 10, 12, and 14-bit are obtained from corresponding RAW images with the image processing pipeline, which shows the necessity of RAW images.
The image processing pipeline includes digital gain, white balance, demosaicing, color correction, and gamma correction.
It is worth noting that quantizing the captured 14-bit RAW images to RAW images of various color encodings can be different from directly capturing them.
However, since a specific commercial camera model only support RAW capture with constant bit depth (typically 14 bits in high-end DSLR), here, we simulate their results with quantization for fast verification.

As shown in Table~\ref{tab:bits}, we can see the performance of both sRGB-dark and RAW-dark gradually increase from 8-bit to 14-bit, and the results of sRGB-dark are very similar to RAW-dark. 
Besides, we also notice that the results of RAW-dark are slightly better than the sRGB-dark.
The reason may be that the steps in the image processing pipeline can make the noise of sRGB more complex than RAW~\citep{brooks2019unprocessing}, which leads to accuracy degradation.
These quantitative results show that the high-bit property of RAW images plays a crucial part in low-light instance segmentation.

\vspace{+1mm}
\noindent\textbf{Low-light Synthetic RAW Pipeline.~}
The unprocessing operation inverts sRGB images to synthetical RAW images, and noise injection simulates the corruption caused by limited photon count and imperfection of photodetectors.
As shown in Table~\ref{tab:abl-syn}, they bring in 2.2 AP and 3.5 AP performance improvements{\color{black}, respectively}.
Moreover, when we combine these two steps together, the accuracy increases from 31.6 AP to 38.0 AP, which is very close to the result of training with real RAW-dark images 39.0 AP.
It shows our synthetic pipeline is able to generate realistic RAW images.
When we adopt the COCO dataset, the trend of results is similar and shows the satisfying generalization ability of low-light RAW synthetic pipeline.

\vspace{+1mm}
\noindent\textbf{Adaptive weighted downsampling (AWD) layer.~}
Table~\ref{tab:downsample} shows AWD reduces the feature disturbance compared with baseline, which means effectiveness on feature denoising.
Though traditional filters can also be helpful, they may blur the foreground signal, which is not optimal.
AWD avoids this problem by predicting content-aware filters. 
It surpasses all traditional low-pass filters and considerably improves the AP by 1.3 and 1.6 points when trained on LIS and COCO{\color{black}, respectively}.
And Table~\ref{tab:gp} verifies the effectiveness of the global pooling branch in the AWD.

To find out the best kernel size of the proposed AWD layer, we conduct experiments on different kernel sizes ranging from $2\times 2$ to $5\times 5$.
As shown in Table~\ref{tab:kernel}, the AWD layer can bring 0.4-1.3 AP improvement, and the kernel size of $3 \times 3$ shows the best results, which is 1.3 AP better than the baseline.

\begin{figure*}[t!]
\centering
\includegraphics[width=0.98\linewidth]{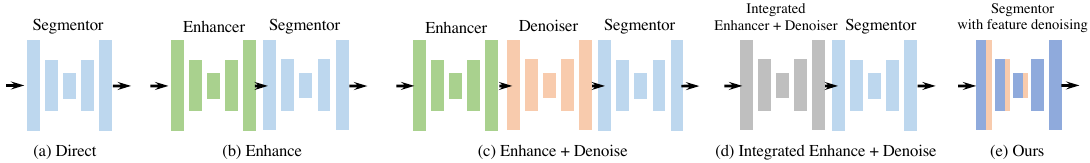} 
\caption{
Competing Methods. 
We illustrate different pipelines for comparison in Table~\ref{tab:main}.
It can be seen that the proposed method is straight and concise.
}
\label{fig:paradigm}
\end{figure*}


\begin{table*}[t!]
\color{black}
\footnotesize
\caption{\small 
Quantitative comparisons of low-light instance segmentation. 
\textbf{
To show the low-light performance in the uncontrolled real world, all models only use COCO data for training and evaluate on test set of the LIS dataset 
}
Our method is trained on synthetic COCO (synthesized by the low-light RAW synthetic pipeline), and evaluated on RAW-dark images in the LIS dataset.
Whereas the Mask R-CNN in other pipelines is optimized with original images in COCO, and evaluated on enhanced sRGB images outputted by the preprocessing methods in the LIS dataset.
}
\label{tab:main}
\centering
\renewcommand{\arraystretch}{1.1}
\scalebox{0.82}{
\begin{tabular}{c|c|c|ccc|ccc|c}
\hline
\multirow{2}{*}{Pipeline}&\multirow{2}{*}{Preprocessing method} & \multirow{2}{*}{Method}&\multirow{2}{*}{AP} & \multirow{2}{*}{AP$_{50}$ } & \multirow{2}{*}{AP$_{75}$ } & \multirow{2}{*}{AP$^{box}$} & \multirow{2}{*}{AP$^{box}_{50}$ } & \multirow{2}{*}{AP$^{box}_{75}$ } & \multirow{2}{*}{FPS}\\
& &  &   &  &   &   &  &  & \\
\hline
Direct & - & Mask R-CNN  
&19.8 &36.2  &18.4 &22.8  &38.4  &24.4  & \bf{56.2}\\
\cline{1-10}
\multirow{5}{*}{Enhance} & HE~\citep{gonzalez2002digital} &Mask R-CNN 
&18.9  &33.0 &18.4 &22.5 &36.5  &24.1 & 42.7 \\
& GLADNet~\citep{wang2018gladnet} &Mask R-CNN 
&14.0 &24.4 &13.8 &16.0 &25.9  &17.5 & 31.2 \\
& Retinex-Net~\citep{wei2018deep} &Mask R-CNN  
&18.2 &30.9 &18.5  &21.0  &33.2 &22.9  & 33.1\\
& EnlightenGAN~\citep{jiang2021enlightengan} &Mask R-CNN  
&19.0 &33.3 &18.6  &22.3 &36.5 &24.2  & 38.0\\
& Zero-DCE~\citep{guo2020zero} &Mask R-CNN  
&19.7 &34.4 &19.1  &22.8 &37.0 &24.2  & 47.4\\
\cline{1-10}
\multirow{5}{*}{Enhance + Denoise} 
& HE~\citep{gonzalez2002digital} + SGN~\citep{gu2019self}   &Mask R-CNN  
&21.0  &36.4 &20.7  &25.1 &40.2 &26.5 & 31.4\\
& GLADNet~\citep{wang2018gladnet} + SGN~\citep{gu2019self}   &Mask R-CNN  
&21.8 &37.5 &21.5 &25.4 &41.0 &26.4 &25.4 \\
& Retinex-Net~\citep{wei2018deep} + SGN~\citep{gu2019self}   &Mask R-CNN  
&22.1  &37.5 &22.2  &25.8 &41.7 &27.8 &26.0 \\
& EnlightenGAN~\citep{jiang2021enlightengan} + SGN~\citep{gu2019self}   &Mask R-CNN  
&26.0 &45.7 &25.0 &30.8 &50.2 &33.1 & 29.0 \\
& Zero-DCE~\citep{guo2020zero} + SGN~\citep{gu2019self}   &Mask R-CNN  
&26.5 &\underline{46.1} &25.9  &\underline{31.2} &\underline{50.4} &\underline{33.9} & 34.1 \\
\cline{1-10}
Integrated &SID~\citep{2018sid} &Mask R-CNN  
&\underline{27.2} &45.4 &\underline{26.5} &30.9 &49.6 & 32.9 &43.8 \\
Enhance + Denoise &REDI~\citep{lamba2021restoring}  &Mask R-CNN  
&23.3  &41.7  &22.3   &27.2  &44.6  &29.0 &44.0 \\
\cline{1-10}
End-to-end (\textbf{Ours})&- &Mask R-CNN &\bf 31.8 &\bf 52.3 &\bf 31.4 &\bf 37.6 &\bf 58.4 &\bf 40.4 & \underline{53.1} \\
\hline
\end{tabular}
}
\end{table*}

{\color{black}
\vspace{+1mm}
\noindent{\bf Comparison with attention mechanisms.}
Here, we compare the AWD layer with attention mechanisms and discuss their essential differences.
First, the AWD layer has a different motivation.
Attention mechanisms are motivated by the human perception that treats information unequally.
They assign different weights to input so as to pay more attention to important information.
While the AWD layer is motivated by the fact that traditional low-pass filters (\eg, Gaussian filter) can suppress high-frequency noise.
To suppress high-frequency feature noise as well as keep the details, the AWD layer is designed to predict input-variant low-pass filters.
Second, the AWD layer is technically different.
Attention mechanisms such as~\citep{2018cbam} aim to improve the convolutional blocks while the AWD layer is proposed to improve the downsampling operation between convolutional blocks, which makes them orthogonal and complementary to each other.
Therefore the AWD layer can steadily further improve the performance of the attention-based model as shown in Table~\ref{tab:attention}.
We compare the proposed AWD with CBAM~\citep{2018cbam} in Table~\ref{tab:attention}.
And we observe two interesting results that are consistent with the analysis.
First, the AWD helps the baseline model achieve better results than CBAM (39.3 AP \emph{vs.} 38.7 AP).
Second, the AWD further improves the performance of CBAM by 1.3 (from 38.7 AP to 40.0 AP), which is the same degree of improvement as the baseline, \ie, the AWD also improves the baseline by 1.3 (from 38.0 AP to 39.3 AP).
These results verify that the improvement brought by AWD is entirely orthogonal to the CBAM.
Furthermore, we conduct more experiments with different attention mechanisms including non-local~\citep{2018nonlocal} and squeeze-and-excitation (SE)~\citep{2018senet}, which are spatial attention and channel attention, respectively.
Their results in Table~\ref{tab:attention} draw the same conclusion, \ie, the AWD helps the baseline model achieve better results than non-local and SE and the AWD can further steadily improve the results of non-local~\citep{2018nonlocal} and SE~\citep{2018senet}.
}

\vspace{+1mm}
\noindent\textbf{Smooth-oriented convolutional block (SCB).}
The SCB explicitly employs a branch to learn to reduce the feature noise with a smooth filter.
And at inference, the SCB can be folded to a normal convolutional layer, which means it boosts the model with no extra computational cost.
Here, we try to replace the smooth-oriented convolution (SConv) with different traditional smooth filters for comparison.
As shown in Table~\ref{tab:scb}, using Gaussian or mean filter also brings performance improvements for low-light instance segmentation.
But due to their fixed filter weight, they fail to learn to deal with the feature noise in a flexible way.
And SConv can be optimized during training, so as to learn the most appropriate filter weights for each channel and achieve better performance.

\begin{table*}[t!]
\color{black}
\footnotesize
\caption{\small Quantitative comparisons of low-light instance segmentation after {\color{black}finetuning} on the LIS dataset.
\textbf{
All methods can access to train set of the LIS dataset for  {\color{black}finetuning} and are evaluated on test set of the LIS dataset.
}
Our method is trained on image pairs of RAW-dark and RAW-normal, and evaluated on RAW-dark images in the LIS dataset.
While the Mask R-CNN in ``Enhance + Denoise'' and ``Integrated Enhance + Denoise'' pipelines use enhanced sRGB/RAW-dark images outputted by the preprocessing methods in the LIS dataset for training/finetuning and evaluation. The {\color{black}U-Net} and Mask R-CNN in``Jointly optimized'' pipeline are jointly optimized and evaluated on RAW-dark images in the LIS dataset, where the U-Net preprocessor is supervised by an extra image restoration ($L2$) loss using RAW-Normal images as ground truth.
}
\label{tab:main_finetune}
\centering
\renewcommand{\arraystretch}{1.1}
\scalebox{0.82}{
\begin{tabular}{c|c|c|ccc|ccc|cc}
\hline
\multirow{2}{*}{Pipeline}&\multirow{2}{*}{Preprocessing method} & \multirow{2}{*}{Method}&\multirow{2}{*}{AP} & \multirow{2}{*}{AP$_{50}$ } & \multirow{2}{*}{AP$_{75}$ } & \multirow{2}{*}{AP$^{box}$} & \multirow{2}{*}{AP$^{box}_{50}$ } & \multirow{2}{*}{AP$^{box}_{75}$ } & \multirow{2}{*}{FPS}\\
& &  &   &  &   &   &  &  \\
\hline
Direct & - & Mask R-CNN &35.5 &57.5 &36.1  & 42.9 & 64.3 & 46.1 & \bf{56.2}\\
Direct (RAW) & - & Mask R-CNN &39.0 &61.3 &40.1 &46.1 &67.8 &50.5 & \bf{56.2}\\
\cline{1-10}
\multirow{2}{*}{Enhance + Denoise} 
& EnlightenGAN~\citep{jiang2021enlightengan} + SGN~\citep{gu2019self}   &Mask R-CNN  &37.1 &60.2 &37.4 &44.5 &67.0 &48.6 & 29.0 \\
& Zero-DCE~\citep{guo2020zero} + SGN~\citep{gu2019self}   &Mask R-CNN  &36.9 &60.3 &37.4  &44.8 &67.5 &49.0 & 34.1 \\
\cline{1-10}
Integrated &SID~\citep{2018sid} &Mask R-CNN  &37.8 &60.0 &38.3 &44.7  &66.6 &46.9 & 43.8 \\
Enhance + Denoise &REDI~\citep{lamba2021restoring}  &Mask R-CNN  &36.0  &59.0  &35.8 &42.8  &66.1  &45.9 & 44.0 \\
\cline{1-10}
Jointly optimized &U-Net~\citep{ronneberger2015u} &Mask R-CNN  &\underline{39.2}  &\underline{61.4}  &\underline{40.0} &\underline{46.2}   &\underline{67.8} &\underline{50.7} & 43.8 \\
\cline{1-10}
End-to-end (\textbf{Ours})&- &Mask R-CNN & \textbf{42.7} & \textbf{66.2} & \textbf{43.3} & \textbf{50.3} & \textbf{72.6} &\textbf{55.2} & \underline{53.1} \\
\hline
\end{tabular}
}
\end{table*}

{\color{black}
\vspace{+1mm}
\noindent\textbf{Disturbance suppression \emph{vs.} Perceptual loss.~}
Here, we compare our disturbance suppression learning with perceptual loss~\citep{2016perceptual, 2020qis} for low-light instance segmentation.
The perceptual loss~\citep{2016perceptual, 2020qis} adopts a teacher-student structure for learning, its pretrained teacher extracts clean features from clean images to supervise the student for noisy images.
Compared with perceptual loss~\citep{2020qis}, the proposed disturbance suppression learning shows two beneficial characteristics.
First, the disturbance suppression learning needs not to pretrain a teacher model, which makes our training simpler and faster.
Second, the disturbance suppression learning can learn discriminative features from both clean and noisy images, whereas ``student" in perceptual loss only sees noisy images and cannot fully utilize the clean images.
Thus the disturbance suppression learning can keep stable accuracy no matter whether images are corrupted by noise or not. 
As shown in Table~\ref{tab:TSL}, on RAW-dark, the disturbance suppression learning increases AP by 0.9 point, while perceptual loss~\citep{2020qis} only brings 0.3 AP improvement.
And on RAW-normal, the disturbance suppression learning shows 1.9 AP improvement while perceptual loss~\citep{2020qis} brings 0.4 AP.
These results verify the above analysis.
}

{\color{black}
\vspace{+1mm}
\noindent\textbf{Extra ablation studies on normal-light dataset.~}
Moreover, we have conducted a series of experiments to evaluate the impact of AWD, SCB, and DSL on the normal-light dataset COCO~\citep{mscoco2014}. 
We train the Mask R-CNN~\citep{MaskRCNN2017} on the COCO~\citep{2014microsoft} for 12 epochs, and the ResNet-50-FPN~\citep{resnet2016, 2017feature} serves as backbone.
As shown in~\ref{tab:coco}, the AWD layer improves the performance by 1.0 AP without bells and whistles.
It means the AWD layer also improves the robustness of networks under normal-light conditions by designing the downsampling process carefully.
As for SCB, it adds a branch to learn smooth filters during training and can be folded back to a normal $3\times3$ convolutional layer by a linear combination.
It helps the convolutional blocks to suppress the high-frequency feature noise caused by image noise.
But the image noise is imperceptible in normal-light images of COCO~\citep{mscoco2014}.
Therefore, the SCB brings minor improvement on COCO~\citep{2014microsoft}.
We further evaluate the DSL for normal light images.
We use the noise injection to synthetic low-light noise images.
Interestingly, though it is proposed for the low-light task, it can also improve performance under normal light.
The possible explanations are i) DSL pushes the model to learn noise-invariant features, which is more discriminative.
ii) DSL makes model learn from both clean images and its noisy version, and the noisy images can be regarded as a kind of augmentation.
}

\vspace{+1mm}
\noindent\textbf{Visualization of learned filter weights.~}
As shown in Figure~\ref{fig:learned_kernel}, we visualize the learned filter weights predicted by the AWD layer. 
It can be seen that the predicted filter weights have a high standard variance for edges of the scene content and a low standard variance for the background.
High variance corresponds to less blur, while low variance corresponds to more blur.
This means the AWD layer can correctly predict content-aware filters to blur high-frequency content (e.g., edges of scene content) less to preserve foreground signals and blur low-frequency background more to suppress the feature noise.

\begin{figure*}[t!]
\vspace{-3mm}
\centering
\includegraphics[width=0.98\linewidth]{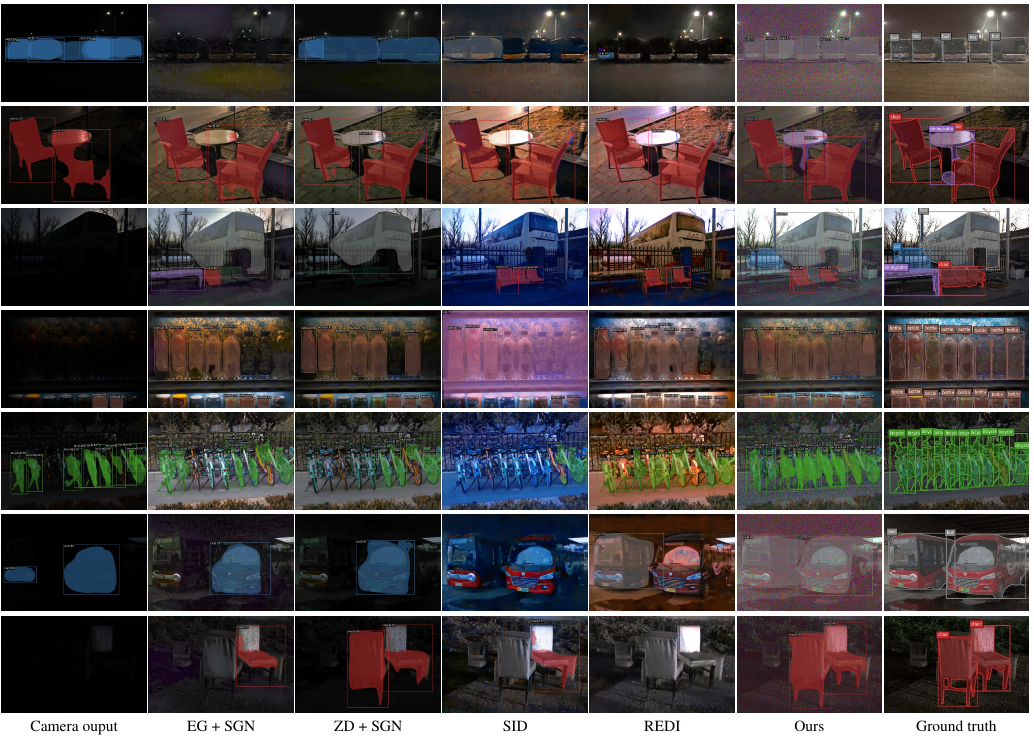}
\caption{
Visual comparisons on our LIS dataset.  
To make it easy to browse, we only show visualized results of the baseline and four of the best competitors. Our RAW results are converted to sRGB for visualization. }
\label{fig:main}
\end{figure*}

\begin{table*}[t!]
\color{black}
\footnotesize
\caption{\small 
Quantitative comparisons of low-light instance segmentation when using Swin-T~\citep{2021swin} and ConvNeXt-T~\citep{2022convnet} as the backbone.
And Mask R-CNN~\citep{MaskRCNN2017} serves as the instance segmentation model.
For brevity, we choose four of the best competitors.
Results are reported on the LIS test set.
The setting for training and testing is the same as Table~\ref{tab:main}.
}
\label{tab:backbone}
\centering
\renewcommand{\arraystretch}{1.1}
\scalebox{0.78}{
\begin{tabular}{c|c|c|ccc|cccc}
\hline
\multirow{2}{*}{Pipeline}&\multirow{2}{*}{Preprocessing Method} & \multirow{2}{*}{Backbone}&\multirow{2}{*}{AP} & \multirow{2}{*}{AP$_{50}$ } & \multirow{2}{*}{AP$_{75}$ } & \multirow{2}{*}{AP$^{box}$} & \multirow{2}{*}{AP$^{box}_{50}$ } & \multirow{2}{*}{AP$^{box}_{75}$ } \\
& &  &   &  &   &   &   & \\
\hline
Direct & - & Swin-T   
&20.7 &38.2  &19.5 &24.8  &41.4  &25.5 \\
\cline{1-9}
\multirow{2}{*}{Enhance + Denoise}
& EnlightenGAN~\citep{jiang2021enlightengan} + SGN~\citep{gu2019self} 
&Swin-T   
&27.8 &48.4 &\underline{28.3} &33.2 &53.9 &\underline{36.3} \\
& Zero-DCE~\citep{guo2020zero} + SGN~\citep{gu2019self} 
&Swin-T  
&\underline{28.3} &\underline{49.7} &27.8  &\underline{33.8} &\underline{54.4} &\underline{36.3}  \\
\cline{1-9}
Integrated 
&SID~\citep{2018sid} 
&Swin-T   
&\underline{28.3} &46.4 &27.5 &31.6 &50.1 &33.0  \\
Enhance + Denoise 
&REDI~\citep{lamba2021restoring} 
&Swin-T   
&25.7  &43.7  &25.3 &29.8 &48.3  &31.2 \\
\cline{1-9}
End-to-end (\textbf{Ours})&- &Swin-T  
& \textbf{32.6} & \textbf{53.5} &\textbf{32.3} & \textbf{37.8} & \textbf{59.2} & \textbf{40.9}  \\
\hline
\hline
Direct & - & ConvNeXt-T   
&23.7 &41.4  &23.7 &27.9  &43.8  &30.0 \\
\cline{1-9}
\multirow{2}{*}{Enhance + Denoise}
& EnlightenGAN~\citep{jiang2021enlightengan} + SGN~\citep{gu2019self} 
&ConvNeXt-T   
&29.5 &50.8 &29.1  &35.9 &\underline{56.2} &39.6 \\
& Zero-DCE~\citep{guo2020zero} + SGN~\citep{gu2019self} 
&ConvNeXt-T  
&30.2 &\underline{51.7} &30.2  &36.1 &\underline{56.2} &\underline{40.0}  \\
\cline{1-9}
Integrated 
&SID~\citep{2018sid} 
&ConvNeXt-T   
&\underline{31.8}  &51.4  &\underline{31.7} &\underline{36.6} &55.1  &\underline{40.0} \\
Enhance + Denoise 
&REDI~\citep{lamba2021restoring} 
&ConvNeXt-T   
&27.6  &46.6  &27.5 &32.2 &49.9  &35.1 \\
\cline{1-9}
End-to-end (\textbf{Ours})&- &ConvNeXt-T  
& \textbf{36.8} & \textbf{58.5} & \textbf{36.9} 
& \textbf{42.7} & \textbf{64.0} &\textbf{47.4} \\
\hline
\end{tabular}
}
\end{table*}

\begin{table*}[t!]
\color{black}
\footnotesize
\caption{\small 
Quantitative comparisons of low-light instance segmentation when using PointRend~\citep{2020pointrend} and Mask2Former~\citep{2022maske2former} as instance segmentation model.
And ResNet-50-FPN~\citep{resnet2016, 2017feature} serves as backbone.
For brevity, we choose four of the best competitors.
Results are reported on the LIS test set.
The setting for training and testing is the same as Table~\ref{tab:main}.
}
\label{tab:pointrend}
\centering
\renewcommand{\arraystretch}{1.1}
\scalebox{0.78}{
\begin{tabular}{c|c|c|ccc|cccc}
\hline
\multirow{2}{*}{Pipeline}&\multirow{2}{*}{Preprocessing method} & \multirow{2}{*}{Method}&\multirow{2}{*}{AP} & \multirow{2}{*}{AP$_{50}$ } & \multirow{2}{*}{AP$_{75}$ } & \multirow{2}{*}{AP$^{box}$} & \multirow{2}{*}{AP$^{box}_{50}$ } & \multirow{2}{*}{AP$^{box}_{75}$ } \\
& &  &   &  &   &   &   & \\
\hline
Direct & - & PointRend  
&20.6 &37.2  &19.2 &23.5  &39.3  &25.1 \\
\cline{1-9}
\multirow{2}{*}{Enhance + Denoise} 
& EnlightenGAN~\citep{jiang2021enlightengan} + SGN~\citep{gu2019self} 
&PointRend  
&26.9 &46.3 &26.2 &31.2 &51.7 &33.3 \\
& Zero-DCE~\citep{guo2020zero} + SGN~\citep{gu2019self}   
&PointRend  
&27.7 &\underline{47.6} &27.3  &\underline{31.9} &\underline{52.0} &\underline{33.7}  \\
\cline{1-9}
Integrated &SID~\citep{2018sid} 
&PointRend  
&\underline{28.3} &46.4 &\underline{27.5} &31.6 &50.1 &33.0  \\
Enhance + Denoise 
&REDI~\citep{lamba2021restoring} 
&PointRend  
&24.0  &42.2  &23.2 &27.7 &46.1  &28.3 \\
\cline{1-9}
End-to-end (\textbf{Ours})&- &PointRend 
& \textbf{32.8} & \textbf{52.9} & \textbf{39.8} & \textbf{37.1} & \textbf{57.9} &\textbf{39.8} \\
\hline
\hline
Direct & - & Mask2Former  
&21.4 &37.9  &20.9 &22.9  &36.9  &23.2 \\
\cline{1-9}
\multirow{2}{*}{Enhance + Denoise} 
& EnlightenGAN~\citep{jiang2021enlightengan} + SGN~\citep{gu2019self} 
&Mask2Former  
&28.0 &47.1 &27.1 &30.9 &48.2 &32.1 \\
& Zero-DCE~\citep{guo2020zero} + SGN~\citep{gu2019self}   
&Mask2Former  
&29.3 &\underline{49.7} &29.1  &31.9 &\underline{50.1} &33.3  \\
\cline{1-9}
Integrated &SID~\citep{2018sid} 
&Mask2Former  
&\underline{31.7} &49.6 &\underline{31.0} &\underline{33.2} &49.5 &\underline{34.3}  \\
Enhance + Denoise 
&REDI~\citep{lamba2021restoring} 
&Mask2Former  
&26.7  &44.1  &26.0 &28.1 &42.9  &29.1 \\
\cline{1-9}
End-to-end (\textbf{Ours})&- &Mask2Former 
& \textbf{35.6} & \textbf{55.2} & \textbf{35.2} & \textbf{37.8} & \textbf{55.9} &\textbf{39.9} \\
\hline
\end{tabular}
}
\end{table*}

\begin{table*}[t!]
\color{black}
\footnotesize
\caption{\small Quantitative comparisons of low-light object detection. 
The backbone is ResNet-50-FPN~\citep{resnet2016, 2017feature}.
Results are reported on the LIS test set.
The setting for training and testing is the same as Table~\ref{tab:main}.
}
\label{tab:detection}
\centering
\renewcommand{\arraystretch}{1.1}
\scalebox{0.98}{
\begin{tabular}{c|c|c|ccc}
\hline
\multirow{2}{*}{Pipeline}&\multirow{2}{*}{Preprocessing method} & \multirow{2}{*}{Method} & \multirow{2}{*}{AP$^{box}$} & \multirow{2}{*}{AP$^{box}_{50}$ } & \multirow{2}{*}{AP$^{box}_{75}$ } \\
& &  &   &  &  \\
\hline
Direct & - & Faster R-CNN  
&21.9 &37.4 &22.4 \\
\cline{1-6}
\multirow{5}{*}{Enhance} & HE~\citep{gonzalez2002digital} &Faster R-CNN  
&22.1 &35.6 &23.5 \\
& GLADNet~\citep{wang2018gladnet} &Faster R-CNN 
&15.4 &24.9 &16.4 \\
& Retinex-Net~\citep{wei2018deep} &Faster R-CNN  
&19.6 &31.1 &21.5 \\
& EnlightenGAN~\citep{jiang2021enlightengan} &Faster R-CNN  
&21.1 &34.8 &21.9 \\
& Zero-DCE~\citep{guo2020zero} &Faster R-CNN  
&22.0 &35.9 &23.5\\
\cline{1-6}
\multirow{5}{*}{Enhance + Denoise} 
& HE~\citep{gonzalez2002digital} + SGN~\citep{gu2019self}   &Faster R-CNN  
&25.1  &39.8 &26.9 \\
& GLADNet~\citep{wang2018gladnet} + SGN~\citep{gu2019self}   &Faster R-CNN  
&24.1 &39.1 &25.3  \\
& Retinex-Net~\citep{wei2018deep} + SGN~\citep{gu2019self}   &Faster R-CNN  
&25.5  &41.0  &27.4 \\
& EnlightenGAN~\citep{jiang2021enlightengan} + SGN~\citep{gu2019self}   &Faster R-CNN  
&29.5  &48.5 & 30.4 \\
& Zero-DCE~\citep{guo2020zero} + SGN~\citep{gu2019self}   &Faster R-CNN  
&\underline{30.5} &\underline{49.8} &\underline{32.5} \\
\cline{1-6}

Integrated &SID~\citep{2018sid} &Faster R-CNN  
& 30.1  &47.6 &32.3 \\
Enhance + Denoise &REDI~\citep{lamba2021restoring}  &Faster R-CNN  
& 29.8  &47.9 &31.6  \\

\cline{1-6}
End-to-end (\textbf{Ours}) &- &Faster R-CNN &\textbf{36.3}  &\textbf{56.6} &\textbf{39.4} \\
\hline
\end{tabular}
}
\end{table*}

\vspace{+1mm}
\noindent\textbf{Visualization of feature maps.~}
As shown in Figure~\ref{fig:featuremap}, we visualize the high-level features of networks.
It can be seen that the proposed method helps to reduce feature disturbance and keep semantic responses to scene content when the image is noisy, which is important for precise low-light instance segmentation.
This visualized result verifies the effectiveness of the proposed method.

\vspace{+1mm}
\noindent\textbf{Summary.~}
To sum up, the RAW images show better potential than sRGB images for the low-light instance segmentation task.
And the low-light synthetic RAW pipeline brings 6.2-6.4 AP improvements when only normal light sRGB images are available (see Table~\ref{tab:abl-syn}).
Furthermore, the proposed method achieves 2.8-3.7 AP improvements to the vanilla model with minor extra computational cost, while replacing ResNet-50-FPN with ResNet-101-FPN only brings 1.5-1.7 AP improvements (see Table~\ref{tab:AWD_SCB_DSL}).
These results substantially demonstrate both the effectiveness and efficiency of the proposed method.

\subsection{Method Comparisons}

In this section, we compare the proposed approach with three types of pipeline, \ie, directly predicting on camera output, predicting on enhanced images, and predicting on enhanced and denoised images.

We select representative traditional (histogram equalization~\citep{gonzalez2002digital}) and learning-based (GLADNet~\citep{wang2018gladnet}, Retinex-Net~\citep{wei2018deep}, EnlightenGAN~\citep{jiang2021enlightengan}, Zero-DCE~\citep{guo2020zero}, SID~\citep{2018sid}, REDI~\citep{lamba2021restoring}) methods as enhancers and adopt the state-of-the-art SGN~\citep{gu2019self} as denoiser.
Considering some competing methods (SID~\citep{2018sid} and REDI~\citep{lamba2021restoring}) already have explicit denoising mechanisms, so we do not append an extra denoising step.
{\color{black} 
Notice that SID~\citep{2018sid}, REDI~\citep{lamba2021restoring}, and the proposed method are designed for taking RAW images as inputs while the rest of methods take sRGB images as inputs.
}
All pipeline is illustrated in Figure~\ref{fig:paradigm}.

For fairness, all {\color{black}settings} use the same instance segmentation model (Mask R-CNN~\citep{MaskRCNN2017}). 
To accurately reflect the practical use and unbiasedly evaluate the proposed method in real-world low-light environments, \textit{we assume the LIS dataset is never seen by any methods during training}, \ie, we regard the whole LIS dataset as a test set.

As shown in Table~\ref{tab:main}, without any preprocessing steps, the baseline normal instance segmentation model only has 19.8 AP, which shows limited accuracy in low-light conditions.
After casting the enhancer to the pipeline, we intuitively expect performance improvement, but the accuracies stay the same (with histogram equalization~\citep{gonzalez2002digital} and Zero-DCE~\citep{guo2020zero}) or even decrease (with GLADNet~\citep{wang2018gladnet}, Retinex-Net~\citep{wei2018deep} and EnlightenGAN~\citep{jiang2021enlightengan}).
We guess the reason is that these enhancers only improve the overall brightness but cannot handle the noise.
To verify it, we further introduce denoiser to the pipeline, and the overall accuracy significantly increases as expected, \eg, Zero-DCE~\citep{guo2020zero} plus SGN~\citep{gu2019self} leads to 6.7 AP gain.
Notice that these methods for comparison use camera outputs.
Then, we also perform experiments with SID~\citep{2018sid} and REDI~\citep{lamba2021restoring}, which can restore sRGB images from low-light RAW images.
And numerical results are surprisingly good, \ie, 27.2 AP with SID~\citep{2018sid}, which outperforms baseline by 7.4 points.
This implies the superiority of using RAW images.

Though these enhancing and denoising steps boost the low-light instance segmentation performance remarkably, our method achieves the best quantitative results without extra preprocessing steps.
Besides, the inference speed of the proposed method outperforms all other pipelines.
And its speed is very close to the original Mask R-CNN~\citep{MaskRCNN2017}.
Moreover, qualitative results illustrated in Figure~\ref{fig:main} show the proposed method can consistently recall most of the targets even in challenging scenarios.

{\color{black}
\subsection{Finetuning on LIS Dataset}
To compare the proposed approach more comprehensively, we choose four of the best pipelines in Table~\ref{tab:main} for further comparison.
Here, all methods can access to train set of the LIS dataset for finetuning, and are evaluated on test set of the LIS dataset.
Moreover, we implement a jointly optimized pipeline (U-Net + Mask R-CNN, the U-Net~\citep{ronneberger2015u} is trained to recover clean normal light images from low-light images) for competition.

The results are shown in Table~\ref{tab:main_finetune}.
It can be seen that all pipelines show better performance with the help of real low-light image pairs in the LIS dataset.
This shows the low-light image dataset is important and necessary for solving the low-light instance segmentation task.
The jointly optimized pipeline outperforms all other pipelines but still underperforms compared with the proposed method by a large margin.
Moreover, our solution shows a much higher inference speed than the jointly optimized and other pipelines.
}

\subsection{Evaluation with Different Instance Segmentation Model}
\label{sec:pointrend}
{\color{black}
We note that the proposed method is model-agnostic, \ie, it should work well with existing methods~\citep{MaskRCNN2017,maskscoringrcnn2019,yolact2019,htc2019,centermask2020, blendmask2020, 2020pointrend,2022maske2former}.
Here, in addition to Mask R-CNN~\citep{MaskRCNN2017}, we perform a series of extra experiments with CNN-based PointRend~\citep{2020pointrend} and recent transformer-based Mask2Former~\citep{2022maske2former}.

Similarly, as shown in Table~\ref{tab:pointrend}, casting enhancers  (EnlightenGAN~\citep{jiang2021enlightengan} and Zero-DCE~\citep{guo2020zero}) and denoiser~\citep{gu2019self} to the pipeline can bring significant improvement.
We also compare our method with SID~\citep{2018sid} and REDI~\citep{lamba2021restoring}, which can restore sRGB images from RAW images.
Though they largely boost performance compared with the baseline, our method achieves the best results.
Notice that our method does not see any real raw images during training and can infer without any preprocessing steps. 
But SID~\citep{2018sid} and REDI~\citep{lamba2021restoring} require extra real-world paired RAW images for training.
}

{\color{black}
\subsection{Evaluation with Different Backbones}
To further verify the proposed method, we also conduct experiments with recent great works of backbones, including the transformer-based Swin Transformer~\citep{2021swin} and CNN-based ConvNeXt~\citep{2022convnet}.

As shown in Table~\ref{tab:backbone}, though the proposed smooth-oriented convolutional block is not available for the transformer-based Swin Transformer~\citep{2021swin}, the proposed method still shows consistent advantages compared with other competitors.
And the results with ConvNeXt~\citep{2022convnet} also draw the same conclusion.
}

\subsection{Extension to Low-Light Object Detection}
\label{sec:llod}
In addition to low-light instance segmentation, the proposed method and dataset can also apply to object detection~\citep{liu2020deep} in the low-light environment.
We perform our experiments with the classic Faster R-CNN~\citep{fasterRCNN2015} detector, and results are provided in Table~\ref{tab:detection}.
We can see that the trend and conclusion are similar to that of low-light instance segmentation.
It shows the effectiveness and generalization of the proposed method.

\section{Conclusion}
{\color{black}
This paper explores end-to-end instance segmentation in very low light on RAW images.
To deal with model degradation in low-light images, we propose adaptive weight downsample layer, smooth-oriented convolutional block, and disturbance suppression learning to handle the feature noise caused by notorious noise in low-light images.
They can reduce feature noise during downsampling and convolution operation, and help the model learn disturbance-invariant features{\color{black}, respectively}.
Noticeably, the proposed method outperforms state-of-the-art competitors by a large margin with less computational cost.

Moreover, we also collect and annotate a large-scale real-world low-light instance segmentation dataset, which contains more than two thousand paired low/normal-light images with instance-level pixel-wise annotations.
It can serve as a benchmark for high-level tasks in low-light conditions. 
We hope that our dataset and the experimental findings can inspire more work on vision in extremely low light in future research. 
}

\acknowledgement
This work was supported by the National Natural Science Foundation of China under Grants No. 62171038, No. 61936011, No. 62088101, and No. 62006023.
Felix Heide was supported by an NSF CAREER Award (2047359), a Packard Foundation Fellowship, a Sloan Research Fellowship, a Sony Young Faculty Award, a Project X Innovation Award, and an Amazon Science Research Award.



%
%


{
\bibliographystyle{spbasic} 
\bibliography{egbib}
}

\end{document}